\documentclass[conference]{IEEEtran}
\usepackage{IEEEtrantools}
\IEEEoverridecommandlockouts

\usepackage{cite}
\usepackage{amsmath,amssymb,amsfonts}
\usepackage{algorithmic}
\usepackage{graphicx}
\usepackage{textcomp}
\usepackage{xcolor}
\usepackage{enumitem}
\usepackage{booktabs}
\usepackage{multirow}
\usepackage{subfigure}
\usepackage{hyperref}
\usepackage{tikz}
\usetikzlibrary{shapes.geometric, arrows.meta, positioning, shadows, patterns, decorations.pathreplacing, shapes, backgrounds}

\def\BibTeX{{\rm B\kern-.05em{\sc i\kern-.025em b}\kern-.08em
    T\kern-.1667em\lower.7ex\hbox{E}\kern-.125emX}}

\title{CoT-X: An Adaptive Framework for Cross-Model Chain-of-Thought Transfer and Optimization}

\author{
\IEEEauthorblockN{
Ziqian Bi$^{*,1}$,
Yinzhi Wang$^{*,2}$,
Tianyang Wang$^{3}$,
Junfeng Hao$^{4}$,\\
Benji Peng$^{5}$,
Wenqian Weng$^{6}$,
Jiayi Gu$^{7}$,
Jacqueline Pang$^{8}$,
Xinyuan Song$^{\dagger,9}$
}
\IEEEauthorblockA{$^{1}$\textit{Purdue University}, USA}
\IEEEauthorblockA{$^{2}$\textit{Baruch College, City University of New York}, USA}
\IEEEauthorblockA{$^{3}$\textit{The Ohio State University}, USA}
\IEEEauthorblockA{$^{4}$\textit{AI Agent Lab}, USA}
\IEEEauthorblockA{$^{5}$\textit{Appcubic}, USA}
\IEEEauthorblockA{$^{6}$\textit{Wayne State University}, USA}
\IEEEauthorblockA{$^{7}$\textit{Central University of Finance and Economics}, China}
\IEEEauthorblockA{$^{8}$\textit{Cornell University}, USA}
\IEEEauthorblockA{$^{9}$\textit{Emory University}, USA}
\thanks{$*$ Equal Contribution, $^{\dagger}$Corresponding author: Xinyuan Song (xsong30@emory.edu)}
}

\date{}

\begin{document}

\maketitle

\begin{abstract}
Long Chain-of-Thought (CoT) traces can improve reasoning accuracy, but repeatedly generating them is costly for smaller or latency-constrained language models. This paper studies a practical alternative: produce a rich rationale once with a capable \emph{thinking} model, compress it, and reuse the compressed trace as context for a cheaper \emph{answering} model. We introduce CoT-X, an adaptive framework for cross-model CoT transfer. CoT-X segments reasoning traces into semantic units, scores their diagnostic and logical importance, selects budget-feasible evidence paths, and reconstructs a coherent compressed rationale for the answering model. On $7,501$ Japanese medical licensing questions spanning $10$ specialties, CoT-X improves accuracy over direct truncation by up to $40.5\%$ under the same token budget, with the largest gains at $64$--$256$ tokens. Across $64$ thinking--answering pairs from eight DeepSeek-R1 and Qwen3 models (1.5B--32B parameters), reasoning transfer is most reliable within a model family, yet remains effective across families once compression normalizes the trace. A Gaussian Process Bayesian optimization layer finds near-optimal model--budget configurations with $15$ evaluations rather than an exhaustive search over all $64$ pairs, reducing evaluation cost by $84\%$. These results show that reasoning quality, token budget, and model compatibility can be optimized jointly, making CoT-style reasoning more practical under realistic deployment constraints.
\end{abstract}


\section{Introduction}

Chain-of-Thought (CoT) prompting has made explicit reasoning a central interface for large language models (LLMs)~\cite{wei2022chain,kojima2023large}. By externalizing intermediate steps, LLMs can solve tasks that require multi-step deduction, mathematical reasoning~\cite{cobbe2021gsm8k,hendrycks2021mmlu,srivastava2023bigbench}, and domain-specific knowledge integration~\cite{singhal2023clinical,jin2021medqa,jin2019pubmedqa,pal2022medmcqa}. This makes CoT attractive for educational tutoring, theorem proving, and medical decision support, where a final option without a rationale is often too brittle to inspect.

Reasoning-specialized models, including DeepSeek-R1~\cite{deepseekr1} and Qwen3~\cite{qwen3}, have amplified both the promise and the cost of this interface. Closed and API-served systems such as GPT-4, Claude 3, and Gemini also show strong reasoning ability~\cite{openai2023gpt4,anthropic2024claude3,geminiteam2023gemini,nori2023capabilities}, but their costs and controllability differ sharply from local open-weight deployment. In medical QA, a useful rationale may need to connect symptoms~\cite{singhal2023clinical,lievin2024medicalreasoning}, differential diagnoses, evidence evaluation~\cite{medcot2024}, and treatment choices~\cite{singhal2023clinical}. Such traces are valuable because they expose the evidential path behind an answer, but they also turn reasoning into a major inference-time bottleneck.

The deployment problem is therefore not whether CoT is useful, but whether its useful information can be delivered within realistic budgets. Long traces increase latency, memory traffic, and serving cost~\cite{yao2023treeofthought,wang2023selfconsistency}; in our medical setting, generated reasoning commonly exceeds the budgets available to smaller answering models. The gap is especially acute in edge and high-throughput serving scenarios, where systems must trade accuracy against response time, memory footprint, and cost per query~\cite{kwon2023vllm,jiang2023llmlingua}.

A natural solution is to separate \emph{reasoning generation} from \emph{answer generation}. A large model can produce a high-quality trace once, while a smaller model reuses that trace to answer cheaply~\cite{magister2023teaching,hsieh2023distilling}. This ``reason once, reuse many times'' paradigm is attractive for cloud--edge systems: the cloud performs an expensive diagnostic analysis, and local or smaller server models consume a cached rationale for low-latency prediction~\cite{kwon2023vllm,jiang2023llmlingua}. If successful, this shifts the bottleneck from repeatedly generating long CoT traces to efficiently representing and transferring them.

The difficulty is that a reasoning trace is not ordinary context. Its value depends on which evidence, intermediate claims, and causal links survive compression. Naive truncation preserves position rather than importance: it may keep setup tokens while discarding the decisive diagnostic step, or cut a rationale before the conclusion is justified. This is closely related to broader challenges in long-context use, prompt compression, and holistic evaluation~\cite{brown2020gpt3,yao2023treeofthought,jiang2023llmlingua,liu2024lostmiddle,liang2023helm}. The central question of this paper is therefore: \emph{Can we compress CoT traces so that smaller models receive the reasoning they need, not merely the first tokens that fit?}

We propose \textbf{CoT-X}, an adaptive framework for cross-model CoT transfer. CoT-X treats the source model as a \emph{thinking model} and the target model as an \emph{answering model}. Rather than training the answering model, CoT-X operates at inference time: it obtains a detailed reasoning trace, compresses it into a budget-feasible rationale, and feeds the compressed trace to the answering model. This design makes the method compatible with off-the-shelf open models and with deployment settings where fine-tuning is unavailable.

The framework has three components. First, it partitions a trace into semantically coherent reasoning units and scores each unit by reasoning depth, knowledge density, logical connectivity, and conclusion relevance. Second, it propagates these scores over a dependency graph using a PageRank-style update~\cite{page1999pagerank}, then selects a budget-feasible evidence path rather than a prefix. Third, it reconstructs the selected units into a coherent compressed rationale, adding minimal bridges and consistency checks so that the answering model receives a readable chain rather than disconnected fragments.

Compression alone does not solve deployment: practitioners must still choose the thinking model, answering model, token budget, and compression strategy. We therefore add a \textbf{Bayesian optimization layer} that models accuracy, robustness, and efficiency as a stochastic function of these choices. A Gaussian Process surrogate with a Matérn kernel~\cite{rasmussen2006gaussian,snoek2012practical,frazier2018tutorial} encodes smoothness across model scale, family, budget, and strategy, while Expected Improvement guides evaluation toward promising configurations.

We evaluate CoT-X on $7,501$ Japanese medical licensing questions across $10$ specialties, a setting related to recent Japanese and biomedical medical-exam benchmarks~\cite{kasai2023japanese,jiang2025jmedbench,liu2025kokushimd}. Using $64$ thinking--answering pairs from DeepSeek-R1 and Qwen3 models, we ask three questions: when does compressed transfer beat truncation, which model pairs transfer reliably, and how much evaluation is needed to find a good deployment point? Adaptive summarization gives the largest gains under tight budgets; family compatibility matters but does not preclude cross-family transfer; and a GP-based search reaches near-optimal configurations with $84\%$ fewer evaluations than exhaustive search. We also observe a consistent relationship between mean accuracy and cross-specialty variation, summarized by a Pareto frontier that helps identify robust deployment regimes~\cite{kaplan2020scaling,hernandez2021scaling}.

\section{Related Work}\label{rw}
\subsection{Chain-of-Thought Prompting and Reasoning Baselines}

Chain-of-thought (CoT) prompting has become a standard baseline for evaluating reasoning in large language models (LLMs)~\cite{wei2022chain,kojima2023large}. Initial work demonstrated that intermediate reasoning steps improve arithmetic, symbolic, and commonsense reasoning~\cite{cobbe2021gsm8k,hendrycks2021mmlu,srivastava2023bigbench}. Later variants add complementary inference-time mechanisms: self-consistency aggregates multiple reasoning paths~\cite{wang2023selfconsistency}, least-to-most prompting decomposes difficult problems into subproblems~\cite{zhou2023leasttomost}, and tree-of-thoughts explores multiple reasoning branches before committing to an answer~\cite{yao2023treeofthought}. These methods are strong baselines, but they usually spend additional tokens or model calls at every query.

Recent reasoning-optimized models have further advanced CoT-style inference. Models such as DeepSeek-R1~\cite{deepseekr1} and Qwen3~\cite{qwen3} are designed to produce detailed reasoning traces, often much longer than ordinary task outputs. These traces can improve complex reasoning, but their length also turns reasoning into a serving bottleneck~\cite{magister2023teaching}. This tension motivates our focus on transferring the useful content of a trace without requiring every downstream model to regenerate it.

\subsection{Benchmarks, API Models, and Medical Evaluation}

General LLM benchmarks such as MMLU, BIG-bench, and HELM established broad evaluation protocols for knowledge, reasoning, calibration, and robustness~\cite{hendrycks2021mmlu,srivastava2023bigbench,liang2023helm}. Medical QA benchmarks, including MedQA, PubMedQA, MedMCQA, and MultiMedQA-style evaluations, further stress clinical knowledge integration and domain-specific reasoning~\cite{jin2021medqa,jin2019pubmedqa,pal2022medmcqa,singhal2023clinical,nori2023capabilities}. Japanese medical licensing benchmarks have recently received separate attention because translation quality, local terminology, and licensing-exam structure can change model behavior~\cite{kasai2023japanese,jiang2025jmedbench,liu2025kokushimd}.

The model ecosystem also matters. API-served models such as GPT-4, Claude 3, and Gemini provide strong reasoning performance but limited control over fine-tuning, decoding internals, and deployment cost~\cite{openai2023gpt4,anthropic2024claude3,geminiteam2023gemini}. Open-weight reasoning models give more control over serving and batching, but still face latency and memory constraints when long CoT traces are generated repeatedly~\cite{deepseekr1,qwen3,kwon2023vllm}. CoT-X is designed for this mixed setting: it does not require model-weight updates, and it can therefore be used with open models, local deployments, or API-style systems whenever a source rationale can be obtained.

\subsection{Model Compression and Knowledge Distillation}

The challenge of deploying large models in resource-constrained environments has motivated extensive research in model compression~\cite{cheng2017survey}. Knowledge distillation~\cite{hinton2015distilling} enables smaller student models to learn from larger teacher models by mimicking their output distributions. This approach has been successfully applied to various architectures~\cite{touvron2021training}, reducing model size while maintaining performance. Recent work has extended distillation concepts to reasoning tasks, with methods such as chain-of-thought distillation~\cite{magister2023teaching}, step-by-step distillation~\cite{hsieh2023distilling}, and rationale-augmented supervision targeting reasoning capability transfer.

However, these approaches typically require training or fine-tuning the student model, which may be infeasible under limited data, restricted compute, or API-only access. CoT-X instead performs inference-time transfer: it changes the reasoning context supplied to the answering model, not the model weights. This makes the method complementary to distillation; a distilled small model could still benefit from compressed traces at inference time.

\subsection{Text Summarization and Information Extraction}

Automatic text summarization has a long history in natural language processing~\cite{nenkova2012survey,see2017get}. Traditional extractive methods select important sentences using features such as term frequency and semantic similarity, while modern abstractive techniques employ transformer-based architectures to generate concise summaries~\cite{lewis2020bart,zhang2020pegasus}. Recent surveys document the shift from neural and pre-trained summarizers toward LLM-based summarization~\cite{zhang2024summarization}, showing their ability to preserve salient information while maintaining coherence.

Our summarization agent builds upon these foundations but addresses the unique challenges of compressing reasoning chains~\cite{cheng2024compressedcot,xia2025tokenskip}. Unlike general text summarization, which seeks conciseness, reasoning chain compression must preserve logical dependencies and causal consistency to ensure validity for downstream reasoning. The sequential nature of reasoning steps introduces additional structural constraints not typically present in standard summarization tasks. Our three-stage pipeline specifically addresses these issues through semantic segmentation, importance propagation, and coherence reconstruction.

\subsection{Prompt Compression and Long-Context Reasoning}

Prompt compression methods reduce inference cost by removing redundant or low-utility tokens before generation~\cite{jiang2023llmlingua}. Long-context studies further show that models do not use all positions equally well; relevant information can be underused when it appears in unfavorable positions or is surrounded by distractors~\cite{liu2024lostmiddle}. These findings are closely related to CoT transfer because a long reasoning trace is a structured context whose usefulness depends on which evidence survives and where it appears.

CoT-X differs from generic prompt compression in both objective and structure. The goal is not merely to shorten a prompt, but to preserve the causal path that supports an answer. We therefore score reasoning segments by diagnostic importance, propagate dependencies across a reasoning graph, and reconstruct a coherent compressed rationale. This makes the compression target task-aware and model-transfer-aware rather than purely token-efficient.

\subsection{Bayesian Optimization for Neural Architecture Search}

Bayesian optimization (BO) has proven effective for hyperparameter tuning and neural architecture search, especially when evaluations are computationally expensive~\cite{snoek2012practical}. It leverages a probabilistic surrogate model, typically a Gaussian process, and an acquisition function to balance exploration and exploitation~\cite{frazier2018tutorial}. BO has been applied to optimize neural network architectures~\cite{kandasamy2018neural,li2017hyperband}, hyperparameter configurations~\cite{shahriari2016taking}, and, increasingly, large language model deployment choices where each evaluation can involve many model calls.

We adapt Bayesian optimization to model selection for chain-of-thought transfer, developing a framework that identifies near-optimal model combinations with minimal evaluations. Our approach considers multiple objectives, including accuracy, robustness, and computational cost. The key innovation lies in designing kernel functions that encode similarity between model configurations and incorporating prior knowledge of model families and scales. This design enables rapid convergence to high-quality configurations, reducing evaluation costs by approximately 84\% compared to exhaustive search.

\section{Methodology}\label{m}

\begin{figure*}[!ht]
  \centering
  \includegraphics[width=\linewidth]{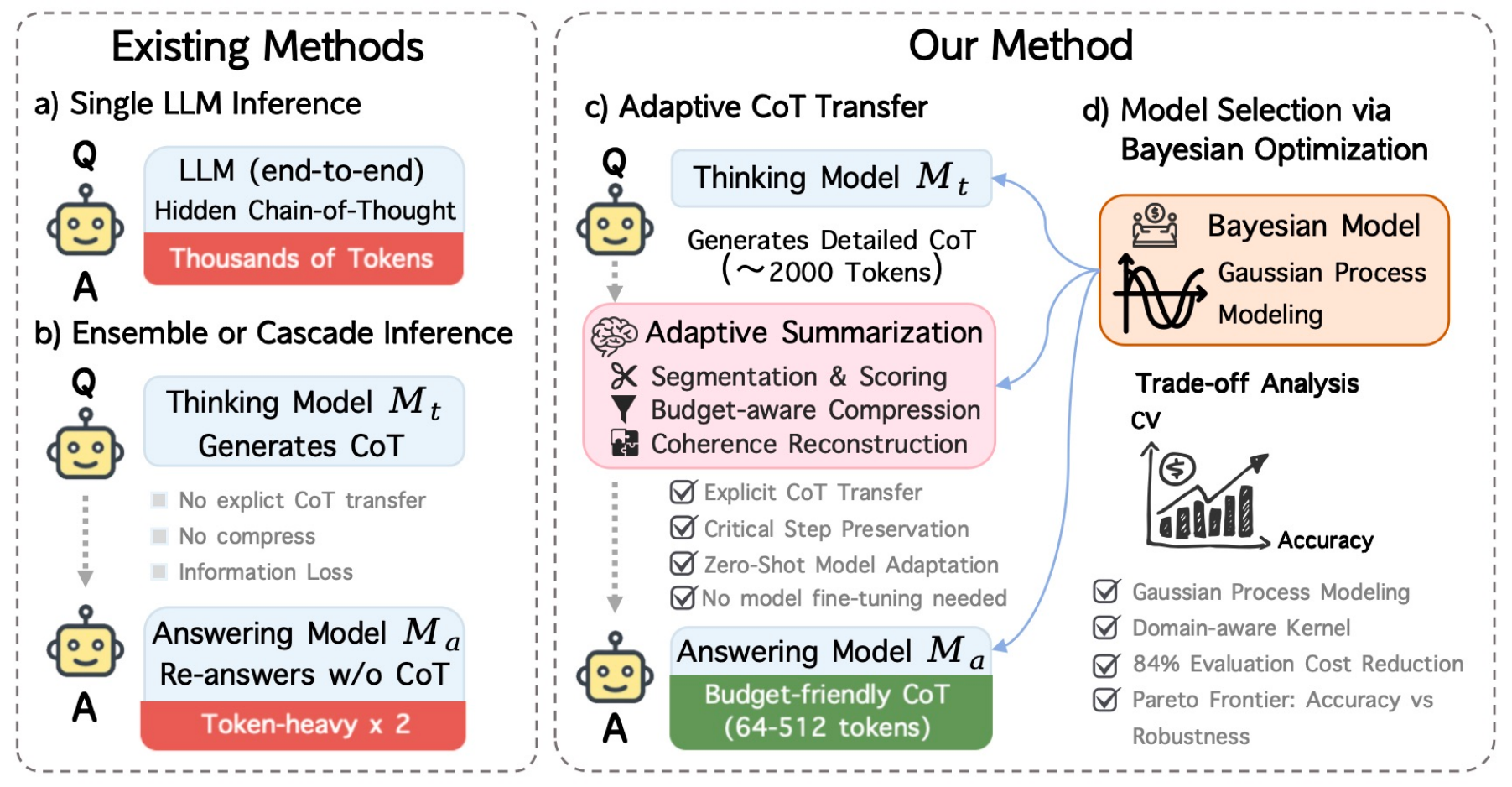}
  \caption{
Inference paradigms compared in this study. 
(a) Single-model reasoning spends the full CoT budget at every query; (b) naive cascades pass only shallow intermediate information and lose much of the reasoning path. 
(c) CoT-X separates a high-capacity thinking model from an efficient answering model by transferring a compressed, budget-aware rationale. 
(d) The Bayesian optimization layer searches over thinking model, answering model, compression strategy, and token budget, reducing the cost of deployment-oriented configuration selection.
  }
  \label{fig:adaptive_cot_pipeline}
\end{figure*}

\begin{figure}[!ht]
    \centering
    \includegraphics[width=\columnwidth]{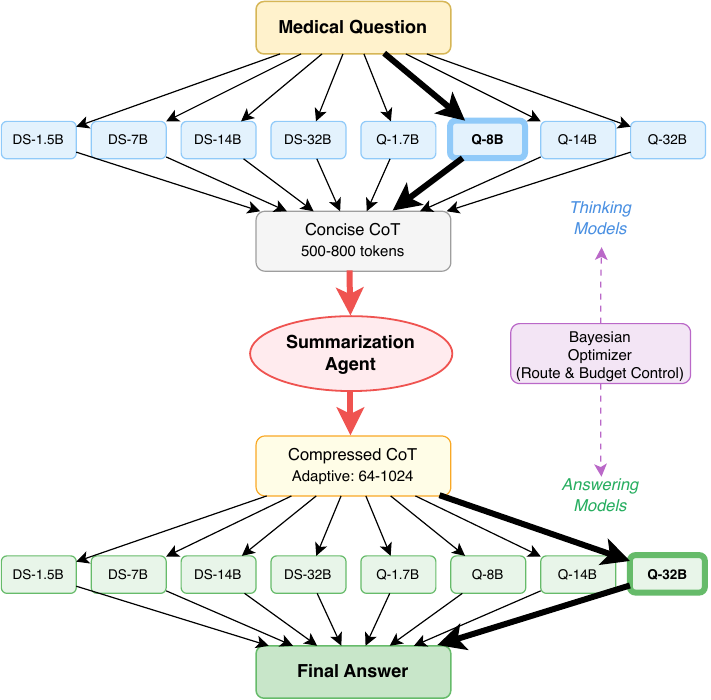}
    \caption{High-level CoT transfer workflow. A thinking model first produces a full reasoning trace; CoT-X compresses the trace into a budget-feasible rationale that preserves diagnostic evidence and conclusion support; the answering model then predicts from the question plus compressed rationale.}
    \label{fig:overview}
\end{figure}

\begin{figure}[!ht]
    \centering
    \includegraphics[width=0.9\columnwidth]{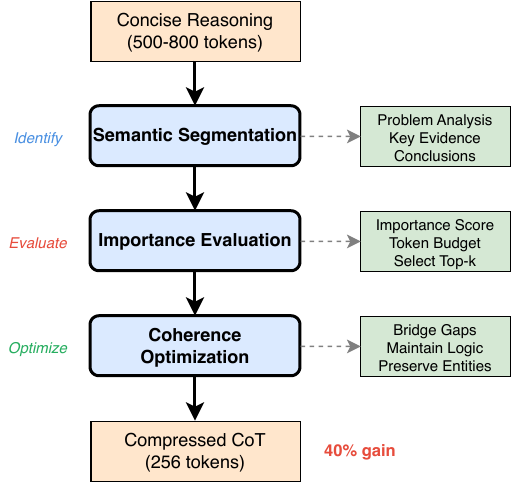}
    \caption{Adaptive compression module. The full trace is segmented into reasoning units, scored by depth, knowledge density, logical connectivity, and conclusion relevance, then selected under the token budget and reconstructed into a fluent rationale. Unlike prefix truncation, the module preserves an evidence path rather than the earliest tokens.}
    \label{fig:pipeline}
\end{figure}

\textbf{Method Overview.} Figure~\ref{fig:adaptive_cot_pipeline} outlines the full inference pipeline. Subfigures (a) and (b) contrast two common baselines: single-model inference, which spends substantial tokens at every query, and cascaded inference without explicit reasoning transfer, which passes only a final prediction or a shallow intermediate representation. CoT-X instead transfers a compressed rationale from a capable thinking model to a cheaper answering model. Figure~\ref{fig:overview} gives the high-level transfer view, while Figure~\ref{fig:pipeline} details the compression module. Together, these components turn long traces into budget-feasible reasoning context; the Bayesian optimization stage in Figure~\ref{fig:adaptive_cot_pipeline}(d) then selects model--budget configurations under an accuracy--robustness trade-off.

\subsection{Problem Formulation}

Let $M_t$ denote a thinking model that produces a detailed reasoning chain, and $M_a$ denote an answering model that generates the final answer given the reasoning context. For a question $q$, the thinking model outputs a reasoning chain $r = M_t(q)$ consisting of $|r|$ tokens. During inference, the answering model processes both the question and the reasoning chain, denoted as $M_a(q, r)$, where $r$ is supplied as additional context for $q$.

A central challenge arises when the token budget $B$ is limited and $|r| > B$. We define a compression function
$f: \mathcal{R} \times \mathbb{N} \rightarrow \mathcal{R}$ that maps a reasoning chain $r$ and a token budget $B$ to a compressed representation $r' = f(r, B)$ such that $|r'| \leq B$ while minimizing the degradation in answer quality.

Formally, the chain-of-thought transfer problem is expressed as:
\begin{equation}
\min_{f} \mathbb{E}_{q \sim \mathcal{Q}} 
\Big[ \mathcal{L}\big(M_a(q, M_t(q)),\, M_a(q, f(M_t(q), B))\big) \Big],
\end{equation}
where $\mathcal{L}$ measures the task-specific loss (e.g., answer accuracy difference or probability divergence), and $\mathcal{Q}$ denotes the question distribution.

The problem involves three key challenges:
\begin{itemize}[left = 0em]
    \item \textbf{Faithfulness}: Retain essential reasoning steps while maintaining logical coherence.
    \item \textbf{Budget Adaptivity}: Flexibly compress reasoning under varying token constraints.
    \item \textbf{Generalization}: Maintain effectiveness across architectures and problem domains.
\end{itemize}

Our adaptive summarization framework addresses these challenges through a multi-stage pipeline that integrates selective extraction, dependency-aware selection, and coherence reconstruction, balancing brevity with informativeness.

\subsection{Adaptive Compression Framework}

The proposed framework starts from the detailed reasoning trace produced by the thinking model and compresses it only after the trace is available. This choice preserves the benefit of a strong reasoner: the source model is allowed to explore differential evidence, intermediate hypotheses, and conclusion support without being forced into a short output format prematurely. The summarization agent then performs budget-aware compression, producing traces at $64$, $128$, $256$, $512$, or $1024$ tokens while preserving the evidence path most useful to the answering model.

For each reasoning segment $s_i$, we compute a composite importance score:
\begin{equation}\label{eq:2}
I(s_i) = \alpha_1 D(s_i) + \alpha_2 K(s_i) + \alpha_3 L(s_i) + \alpha_4 C(s_i),
\end{equation}
where $D(s_i)$ represents reasoning depth (number of inference steps), $K(s_i)$ measures knowledge density (domain-specific term frequency), $L(s_i)$ captures logical connectivity (dependencies between segments), and $C(s_i)$ quantifies conclusion relevance (proximity to the final answer).

Weights $\alpha_1$–$\alpha_4$ are set heuristically to reflect design intuition. Reasoning depth receives higher priority ($\alpha_1 = 0.3$), followed by knowledge density ($\alpha_2 = 0.2$), with logical connectivity and conclusion relevance equally weighted ($\alpha_3 = \alpha_4 = 0.25$). This scheme emphasizes preservation of complete inference chains while maintaining balanced domain and conclusion coverage.

\subsubsection{Importance Propagation and Selection}

Dynamic compression is guided by a dependency graph $G = (V, E)$, where nodes represent reasoning segments and edges represent logical dependencies. Importance scores are propagated using a modified PageRank formulation:
\begin{equation}\label{eq:3}
I'(s_i) = (1-d) + d \sum_{s_j \in \text{pred}(s_i)} \frac{I'(s_j)}{|\text{succ}(s_j)|},
\end{equation}
where $d$ is a damping factor (set to 0.85), and $\text{pred}(s_i)$, $\text{succ}(s_i)$ denote predecessor and successor segments.

A greedy selection algorithm identifies the subset $S^*$ of segments that maximizes total propagated importance within the budget constraint:
\begin{equation}
S^* = \arg\max_{S \subseteq \{s_1, ..., s_n\}} \sum_{s_i \in S} I'(s_i)
\quad \text{s.t.} \quad \sum_{s_i \in S} |s_i| \leq B.
\end{equation}

Compression adapts to available budget $B$. At 64 tokens, only the conclusion and key evidence (top 5\% of segments) are retained. At 128 tokens, the primary reasoning path (top 15\%) is preserved. Budgets of 256, 512, and 1024 tokens include approximately the top 30\%, 50\%, and 75\% of segments, respectively, achieving a balance between coverage and brevity.

\subsubsection{Coherence Reconstruction and Optimization}

To maintain logical coherence, we reconstruct the compressed chain through controlled text generation:
\begin{itemize}[left = 0em]
    \item \textbf{Logical Flow Preservation}: Identify gaps introduced by segment removal and generate concise bridging statements using rule-based templates augmented by LLM refinement~\cite{yao2023treeofthought}.
    \item \textbf{Entity and Relationship Consistency}: Maintain an entity registry to ensure all critical terms and relationships remain contextually defined and consistent, addressing known faithfulness risks in abstractive summarization~\cite{maynez2020faithfulness}.
    \item \textbf{Conclusion Validity}: Validate that the compressed reasoning still supports the final conclusion. Missing evidence triggers either inclusion of additional segments or generation of minimal summary statements~\cite{wang2023selfconsistency}.
\end{itemize}

The algorithm orders selected segments by their original positions, detects logical discontinuities, inserts bridging text for gaps exceeding a threshold, verifies entity integrity, and ensures that the final reasoning leads coherently to the conclusion. This ensures both informational retention and narrative consistency within budget constraints.

\subsection{Bayesian Optimization for Model Selection}

To efficiently identify optimal model configurations, we employ Gaussian Process (GP)–based Bayesian optimization~\cite{snoek2012practical}. The performance function follows:
\begin{equation}
f(x) \sim \mathcal{GP}(\mu(x), k(x, x')),
\end{equation}
where $x$ encodes a configuration (thinking model, answering model, token budget, compression strategy), and $k(x, x')$ is a Matérn kernel:
\begin{align}
k(x, x') &= \sigma^2 \exp\!\left(-\sqrt{5}\frac{d(x,x')}{\ell}\right)
\left(1 + \frac{\sqrt{5}d(x,x')}{\ell} + \frac{5d^2(x,x')}{3\ell^2}\right),
\end{align}
where $d(x,x')$ accounts for model-family similarity, log-scale parameter differences, token-budget ratios, and compression-strategy compatibility.

Expected Improvement (EI) is used as the acquisition function:
\begin{equation}
\text{EI}(x) = \mathbb{E}[\max(0, f(x) - f^*)] = (\mu(x) - f^*)\Phi(Z) + \sigma(x)\phi(Z),
\end{equation}
where $f^*$ is the best observed performance, $Z = (\mu(x) - f^*)/\sigma(x)$, and $\Phi$, $\phi$ denote the standard normal CDF and PDF.

The optimization begins with 8–10 diverse initial samples (smallest/largest models, cross-family and balanced settings), followed by iterative refinement selecting configurations maximizing EI. The process terminates when EI drops below a threshold or the evaluation budget is exhausted, achieving near-optimal performance with about 84\% fewer evaluations than exhaustive search.

\subsection{Performance–Robustness Trade-off Analysis}

We quantify the relationship between average performance and cross-domain robustness using the coefficient of variation (CV):
\begin{equation}
\text{CV} = \frac{\sigma_{\text{domains}}}{\mu_{\text{domains}}},
\end{equation}
where $\sigma_{\text{domains}}$ and $\mu_{\text{domains}}$ are the standard deviation and mean accuracy across medical specialties.

Empirically, we observe a power-law relationship:
\begin{equation}
\text{CV} = \alpha \cdot \text{Acc}^{\beta},
\end{equation}
identified by fitting Pareto-optimal configurations in the performance–robustness space via log-linear regression~\cite{deb2002nsga2},
$\log(\text{CV}) = \log(\alpha) + \beta \cdot \log(\text{Acc})$,
with bootstrap-based uncertainty estimation.

The Pareto frontier is defined as:
\begin{align}
\mathcal{P}^* = \{m \in \mathcal{M} : &\nexists m' \in \mathcal{M}, \nonumber \\
&\text{Acc}(m') > \text{Acc}(m) \land \text{CV}(m') < \text{CV}(m)\}.
\end{align}

We derive two characteristic curves: the Pareto Frontier Curve, representing theoretically optimal trade-offs, and the Typical Performance Curve, capturing practically attainable performance (75th percentile). The area between them defines the feasible solution space. Empirical fitting yields $\alpha \approx 0.42$ and $\beta \approx -2.3$, confirming that accuracy improvements generally reduce cross-domain stability following a predictable power-law scaling.

\section{Experimental Setup}\label{e}
\subsection{Dataset}

We evaluate CoT-X on $7,501$ multiple-choice questions from Japanese national medical licensing examinations. This setting is closely related to recent medical and Japanese-language evaluation efforts, including MedQA-style medical reasoning, PubMedQA, MedMCQA, Japanese medical licensing evaluations, JMedBench, and KokushiMD-10~\cite{jin2021medqa,jin2019pubmedqa,pal2022medmcqa,kasai2023japanese,jiang2025jmedbench,liu2025kokushimd}. It is well suited for CoT transfer because many questions require more than recalling a fact: models must connect clinical cues, domain knowledge, and elimination among plausible answer options. The ten-specialty distribution is reported in Table~\ref{tab:dataset}.

\begin{table}[!ht]
\centering
\caption{Evaluation dataset by specialty. The benchmark contains $7,501$ five-option medical licensing questions across ten Japanese healthcare specialties, creating both large high-coverage subsets (Medicine and Pharmacy) and smaller stress-test subsets (Midwifery, Nursing, and Public Health Nursing) for cross-domain robustness analysis.}
\label{tab:dataset}
\begin{tabular}{lcc}
\toprule
\hline
\textbf{Medical Specialty} & \textbf{Questions} & \textbf{Percentage} \\
\midrule
Medicine & 1,412 & 18.8\% \\
Pharmacy & 1,384 & 18.5\% \\
Dentistry & 974 & 13.0\% \\
Occupational Therapy & 877 & 11.7\% \\
Physical Therapy & 833 & 11.1\% \\
Radiologic Technology & 809 & 10.8\% \\
Optometry & 645 & 8.6\% \\
Midwifery & 206 & 2.7\% \\
Nursing & 183 & 2.4\% \\
Public Health Nursing & 178 & 2.4\% \\
\midrule
\textbf{Total} & \textbf{7,501} & \textbf{100.0\%} \\
\hline
\bottomrule
\end{tabular}
\end{table}

Medicine and Pharmacy form the largest subsets, while Midwifery, Nursing, and Public Health Nursing provide smaller but clinically distinct domains. This imbalance is useful for our study: beyond average accuracy, it allows us to ask whether a transfer configuration remains stable across specialties with different question styles and knowledge demands~\cite{jin2021medqa,singhal2023clinical}. Each item contains a clinical or conceptual prompt, five candidate answers (A--E), and a single correct option. The questions cover differential diagnosis, treatment selection, pharmacological reasoning, procedural knowledge, and public-health judgment~\cite{singhal2023clinical,lievin2024medicalreasoning,medcot2024}.

\subsection{Models}

We evaluate eight open-weight LLMs from two model families: DeepSeek-R1~\cite{deepseekr1} and Qwen3~\cite{qwen3}. DeepSeek-R1 contributes 1.5B, 7B, 14B, and 32B variants; Qwen3 contributes 1.7B, 8B, 14B, and 32B variants. This design gives matched small, medium, and large scales while preserving architectural diversity. Although closed API models such as GPT-4, Claude 3, and Gemini provide important reference points for reasoning capability~\cite{openai2023gpt4,anthropic2024claude3,geminiteam2023gemini}, our grid focuses on open-weight models so that throughput, memory footprint, and cross-model transfer can be measured directly under a controlled serving stack.

Every model is evaluated in both roles: as a thinking model that generates the source rationale and as an answering model that predicts the final option from the question plus compressed reasoning. This yields $8 \times 8 = 64$ transfer pairs. The resulting grid separates scale effects from family effects, allowing us to compare intra-family transfer (e.g., DeepSeek-R1$\rightarrow$DeepSeek-R1) with cross-family transfer (e.g., DeepSeek-R1$\rightarrow$Qwen3)~\cite{magister2023teaching,hsieh2023distilling,kwon2023vllm}.

We compare CoT-X against direct truncation, which keeps the prefix of the source rationale under the same token budget, and against self-performance settings where the thinking and answering roles are played by the same model. These baselines follow the practical alternatives a deployment system would otherwise use: either spend the full reasoning trace on a single model, or cut the trace to fit the answering model's context budget~\cite{wei2022chain,wang2023selfconsistency,jiang2023llmlingua,liu2024lostmiddle}.

\subsection{Implementation Details}

All experiments were run on an 8$\times$NVIDIA H100 cluster under a fixed inference stack. We use \texttt{vLLM}~\cite{kwon2023vllm}, including PagedAttention for KV-cache management, continuous batching, and tensor parallelism for larger models.

We set the maximum sequence length to $4096$ tokens and adapt batch size to model memory footprint. Thinking models use temperature $0.7$ to elicit richer rationales, while answering models use temperature $0.1$ for stable option selection; top-$p$ is fixed at $0.95$. GPU memory utilization is capped at $90\%$. The summarization agent is Qwen3-32B~\cite{qwen3} with temperature $0.3$ and no explicit CoT generation, so compression behaves as a deterministic preprocessing step rather than a second reasoning model.

\subsection{Evaluation Metrics}

We evaluate three dimensions. First, predictive performance is measured by answer accuracy under each token budget and compression strategy. Second, budget efficiency is measured by token efficiency and compression ratio, capturing how much task performance is retained per reasoning token~\cite{magister2023teaching,hsieh2023distilling}. Third, robustness is measured by the coefficient of variation (CV), worst-specialty accuracy, and performance range across medical specialties, following the broader emphasis on multi-metric language-model evaluation~\cite{kaplan2020scaling,hernandez2021scaling,liang2023helm}. For deployment analysis, we also report throughput, latency, and peak GPU memory~\cite{kwon2023vllm}.

For statistical validation, we use paired $t$-tests with Bonferroni correction, $95\%$ bootstrap confidence intervals, and Cohen's $d$ for effect size estimation~\cite{efron1994bootstrap,cohen1988statistical}. Power-law fits are estimated in log space for the performance--robustness analysis. Unless otherwise specified, reported means aggregate over questions within each specialty and then over transfer configurations, with standard deviations used to expose variability rather than hide it.

\section{Results}\label{r}
\subsection{Overall Transfer Behavior}

We first ask whether reasoning traces can be transferred across models at all, and where transfer breaks down. Figure~\ref{fig:performance_matrix_full} reports the complete matrix over $64$ thinking--answering pairs, token budgets, and compression strategies. The dominant pattern is not a single best model, but a structured transfer landscape: same-family pairs form stronger diagonal blocks, large thinking models improve many downstream answerers, and the smallest answerers remain sensitive to the quality and format of the transferred trace.

\begin{figure}[!ht]
\centering
\includegraphics[width=\columnwidth]{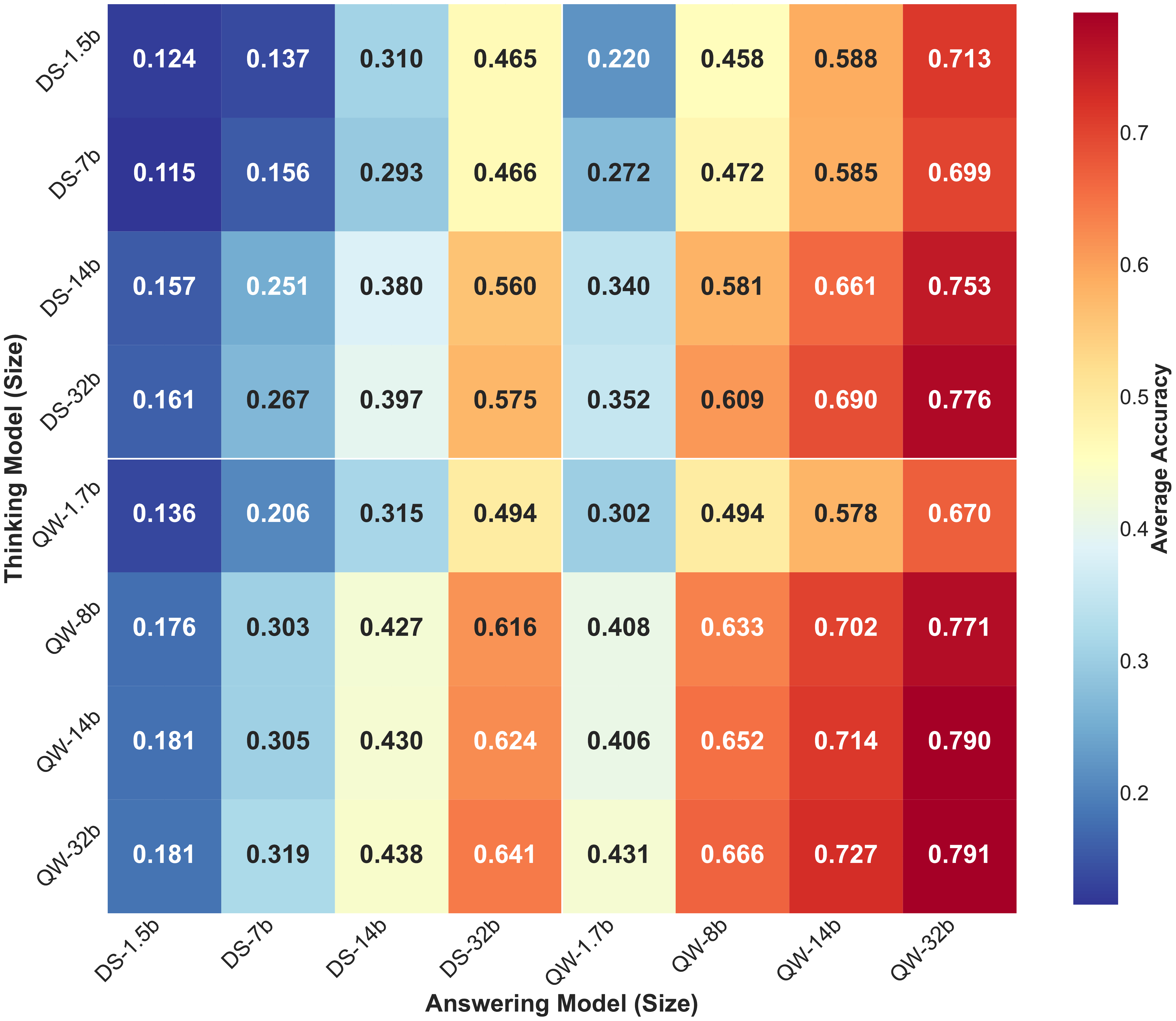}
\caption{Model-pair transfer matrix over the eight DeepSeek-R1 and Qwen3 variants. Rows correspond to thinking models and columns to answering models; block boundaries separate same-family and cross-family transfer. The diagonal and same-family blocks are strongest, while several asymmetric large-to-medium pairings remain competitive, showing that high-quality reasoning can be reused without always deploying the largest model twice.}
\label{fig:transfer_matrix}
\end{figure}

Figure~\ref{fig:transfer_matrix} isolates this structure at the model-pair level. The white boundaries divide the matrix into four transfer regimes: DeepSeek-to-DeepSeek, DeepSeek-to-Qwen, Qwen-to-DeepSeek, and Qwen-to-Qwen. Intra-family transfer is consistently stronger, suggesting that models from the same family produce and consume reasoning in more compatible formats. Cross-family transfer remains viable, but it is more asymmetric: DeepSeek-R1 traces tend to transfer better into Qwen3 answerers than Qwen3 traces transfer into DeepSeek-R1 answerers. This asymmetry is important for deployment, because the best thinking model is not necessarily the best answering model.

Scale produces a second, orthogonal effect. Larger thinking models generate traces that are more useful to many answerers, while larger answering models are better able to exploit compressed context. However, the best efficiency points are often asymmetric. For example, a 32B thinking model paired with a 7B--14B answering model can retain much of the accuracy of a 32B--32B pair while reducing answering-time cost. This motivates the optimization problem studied later: CoT transfer is not simply ``use the largest model twice,'' but a search over compatible and cost-aware pairings.

\subsection{Token Budget and Compression Strategy}

\begin{figure}[!ht]
\centering
\includegraphics[width=\columnwidth]{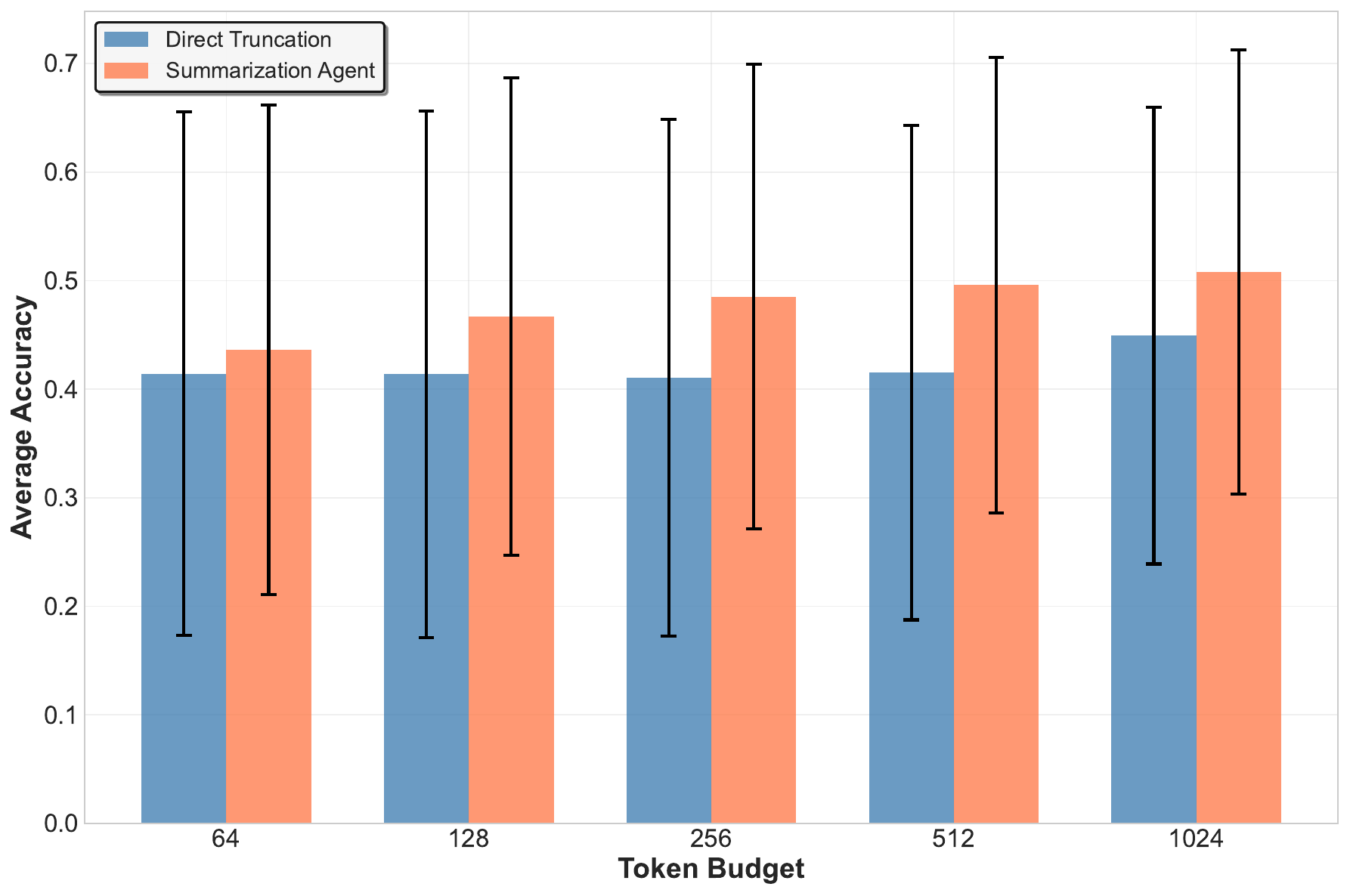}
\caption{Adaptive summarization versus direct truncation under matched token budgets. Bars report average accuracy across transfer configurations, and error bars show standard deviation. CoT-X is strongest in the $64$--$256$ token regime, where direct truncation often removes decisive evidence while adaptive summarization keeps the clinically relevant reasoning path.}
\label{fig:strategy_comparison}
\end{figure}

The central hypothesis of CoT-X is that what matters under a budget is not the length of the retained prefix, but the reasoning content selected for transfer. Figure~\ref{fig:strategy_comparison} supports this hypothesis. At $64$ tokens, adaptive summarization reaches $0.52$ average accuracy compared with $0.37$ for direct truncation, a $40.5\%$ relative improvement. The gain remains substantial at $128$ and $256$ tokens, then gradually narrows as the budget becomes large enough for truncation to preserve more of the original trace.

The error bars tell the same story from a robustness perspective. Summarization reduces variance across model pairs, especially in the low-budget regime where truncation is most likely to remove decisive evidence. In other words, compression improves not only the mean result but also the reliability of reasoning transfer under tight budgets.

\begin{figure*}[!ht]
\centering
\includegraphics[width=\textwidth]{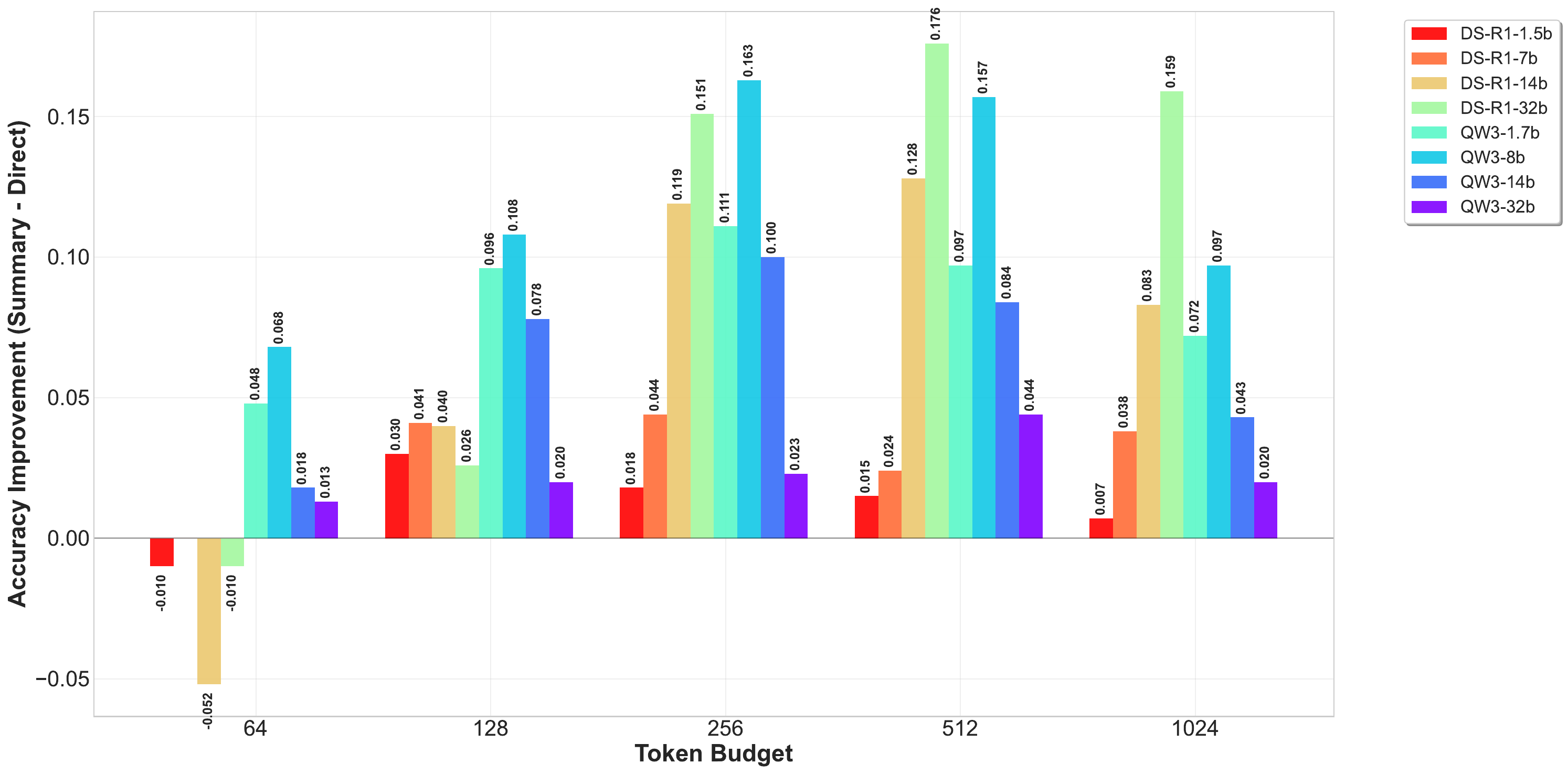}
\caption{Accuracy gain of adaptive summarization over direct truncation for every model pair and token budget. Positive regions indicate that CoT-X preserves more useful reasoning than prefix-based truncation. The largest gains appear for small answerers and cross-family transfers, where the answering model is most sensitive to missing or poorly ordered reasoning evidence.}
\label{fig:improvement_heatmap}
\end{figure*}

Figure~\ref{fig:improvement_heatmap} shows where these gains come from. The largest improvements occur when the answering model is small or when the transfer crosses model families. Both cases are sensitive to presentation: small models have limited capacity to infer missing links from a damaged rationale, while cross-family answerers benefit when the summarizer normalizes the source trace into a more model-agnostic form. The most consistent region is $128$--$256$ tokens, where the budget is large enough to express a complete evidence path but still small enough that selection matters.

\begin{figure}[!ht]
\centering
\includegraphics[width=\columnwidth]{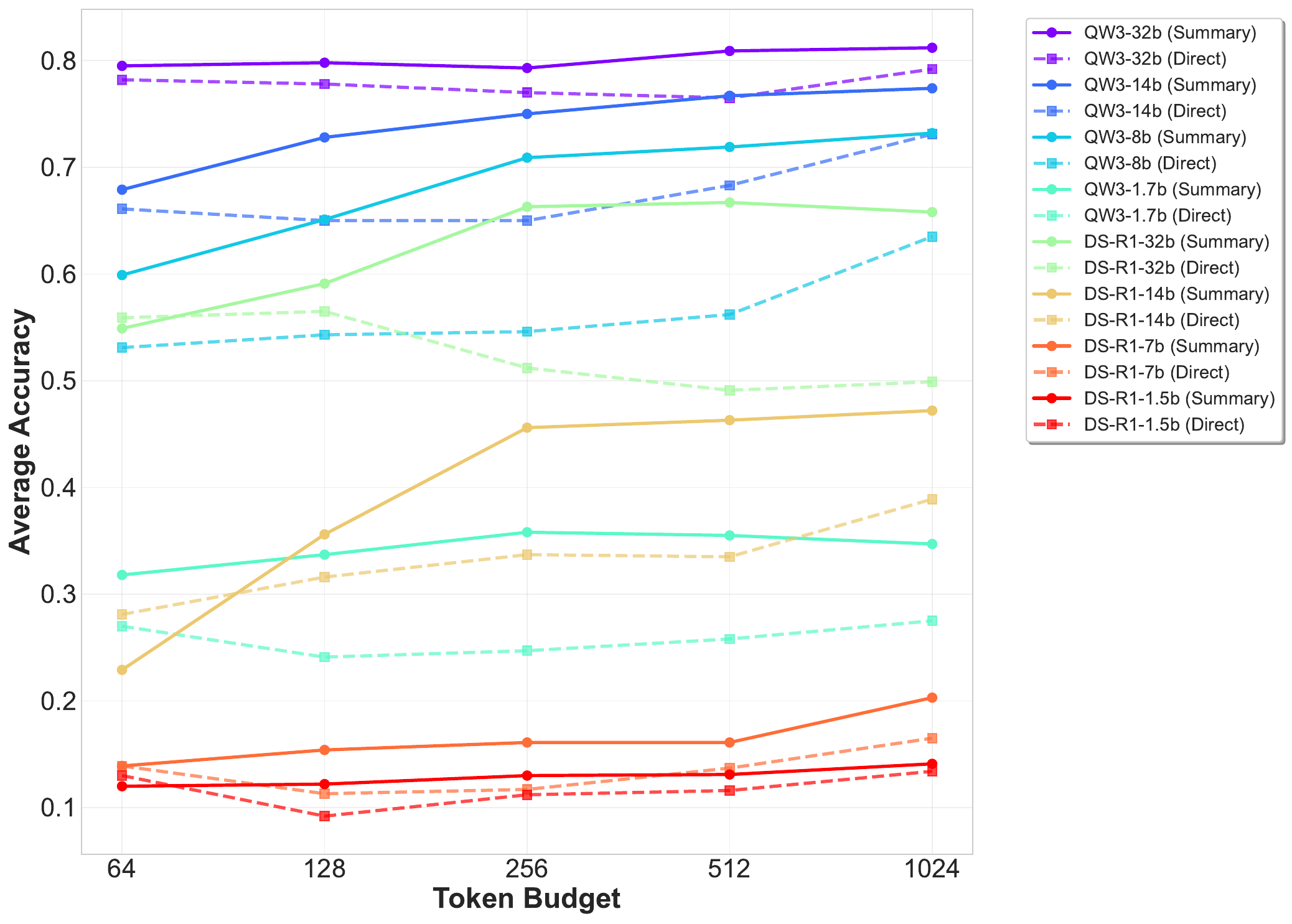}
\caption{Efficiency curves for all $64$ transfer configurations across token budgets. Solid lines denote CoT-X summarization and dashed lines denote direct truncation. CoT-X reaches the accuracy plateau earlier, especially around $256$--$512$ tokens, indicating that adaptive selection converts a limited reasoning budget into useful evidence more efficiently.}
\label{fig:efficiency_curves}
\end{figure}

The efficiency curves in Figure~\ref{fig:efficiency_curves} explain why the mid-budget regime is practically important. Most curves rise sharply from $64$ to $256$ tokens and then flatten around $512$ tokens. This indicates an information saturation point: once the main evidence chain is present, extra tokens mostly add redundant explanation. The solid summarization curves reach this saturation earlier than dashed truncation curves, showing that CoT-X converts token budget into useful reasoning more efficiently.

\begin{figure}[!ht]
\centering
\includegraphics[width=\columnwidth]{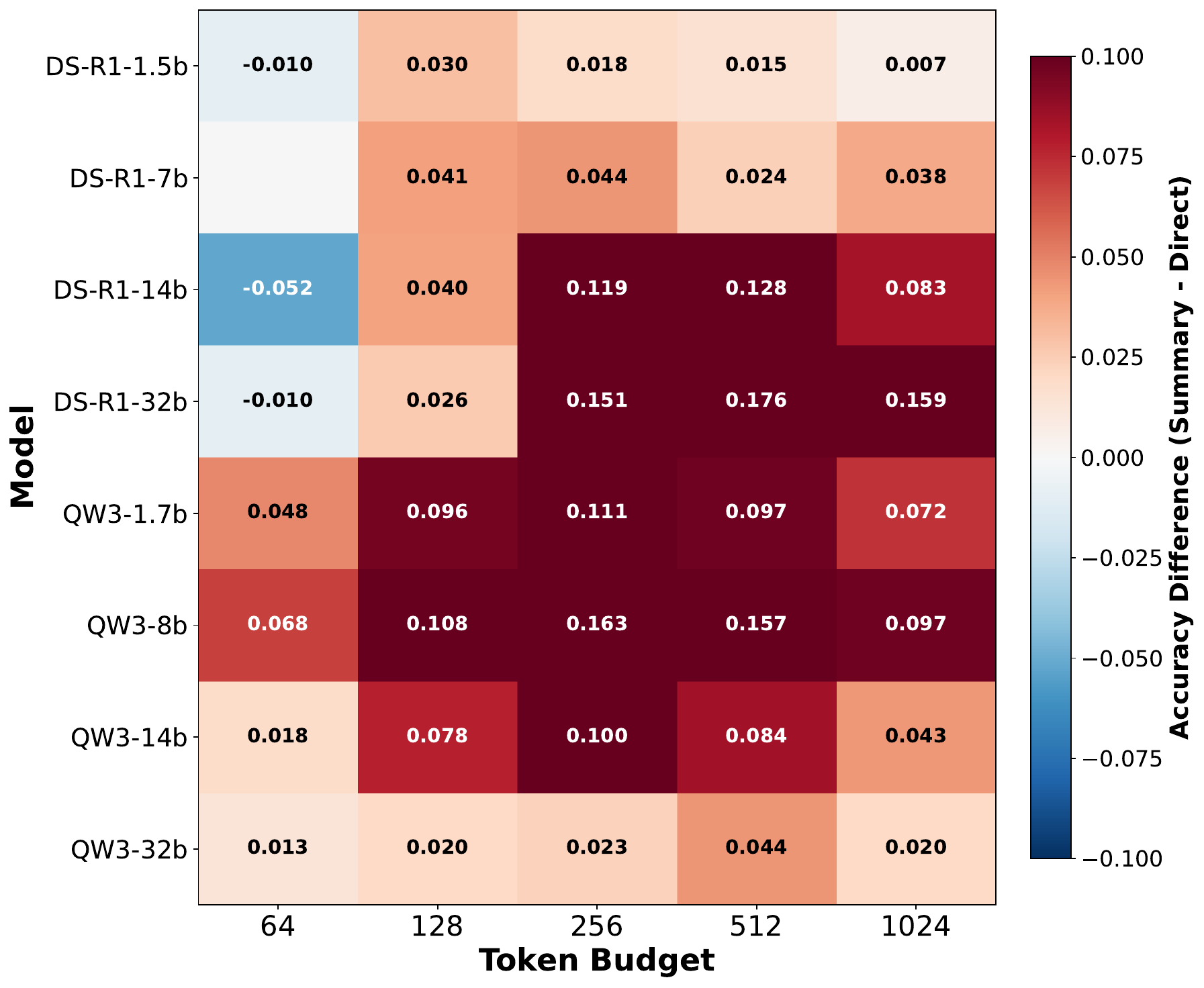}
\caption{Strategy preference by model scale and token budget. Red indicates that adaptive summarization outperforms direct truncation; stronger red indicates a larger margin. The absence of blue regions shows that truncation is not the preferred strategy in this grid, while the strongest red cells identify the deployment settings where CoT-X matters most.}
\label{fig:model_preferences}
\end{figure}

Figure~\ref{fig:model_preferences} summarizes the decision boundary between strategies. The plot contains no region where truncation is preferred, but the magnitude of the summarization advantage depends on budget and scale. The advantage is strongest below $256$ tokens and for smaller answerers; it becomes modest for the largest models at $1024$ tokens. Thus, the safe practical rule is not that summarization is always equally valuable, but that its value is highest exactly where deployment constraints are most severe.

\subsection{Cross-Domain Robustness}

\begin{figure}[!ht]
\centering
\includegraphics[width=\columnwidth]{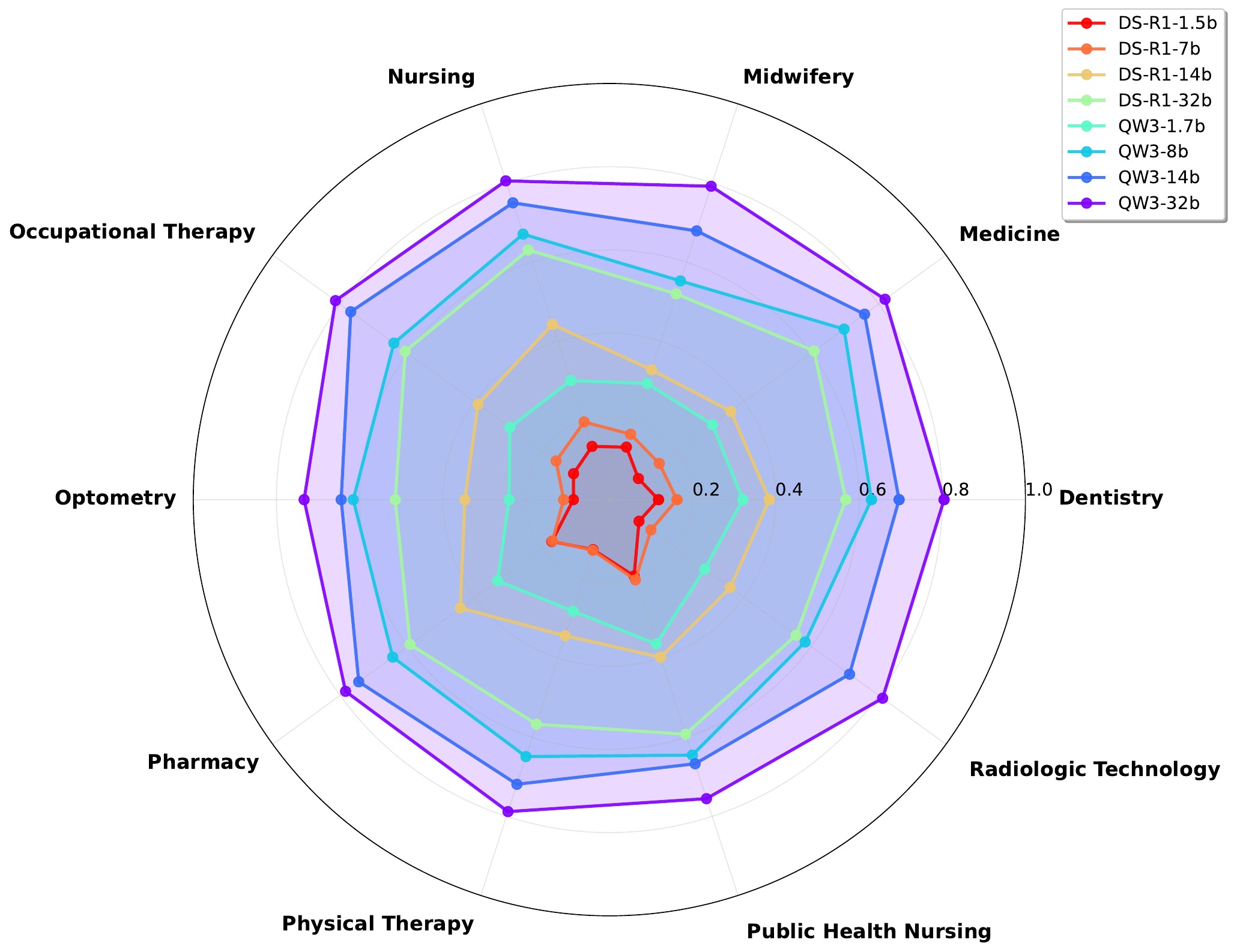}
\caption{Specialty-level accuracy profile for the eight evaluated models. Each axis is one medical specialty, and distance from the center indicates higher accuracy. Larger models show smoother profiles across specialties, while smaller models vary more sharply, motivating the robustness analysis beyond aggregate accuracy.}
\label{fig:radar_chart}
\end{figure}

Average accuracy can hide failure modes concentrated in particular medical specialties. Figure~\ref{fig:radar_chart} therefore breaks performance down by domain. Physical Therapy and Optometry are comparatively easier, while Medicine and Pharmacy are harder, reflecting longer prompts, denser terminology, and more complex differential reasoning. Larger models form smoother profiles across axes, whereas smaller models show sharper peaks and troughs, indicating greater sensitivity to specialty-specific phrasing and knowledge requirements.

\begin{figure}[!ht]
\centering
\includegraphics[width=\columnwidth]{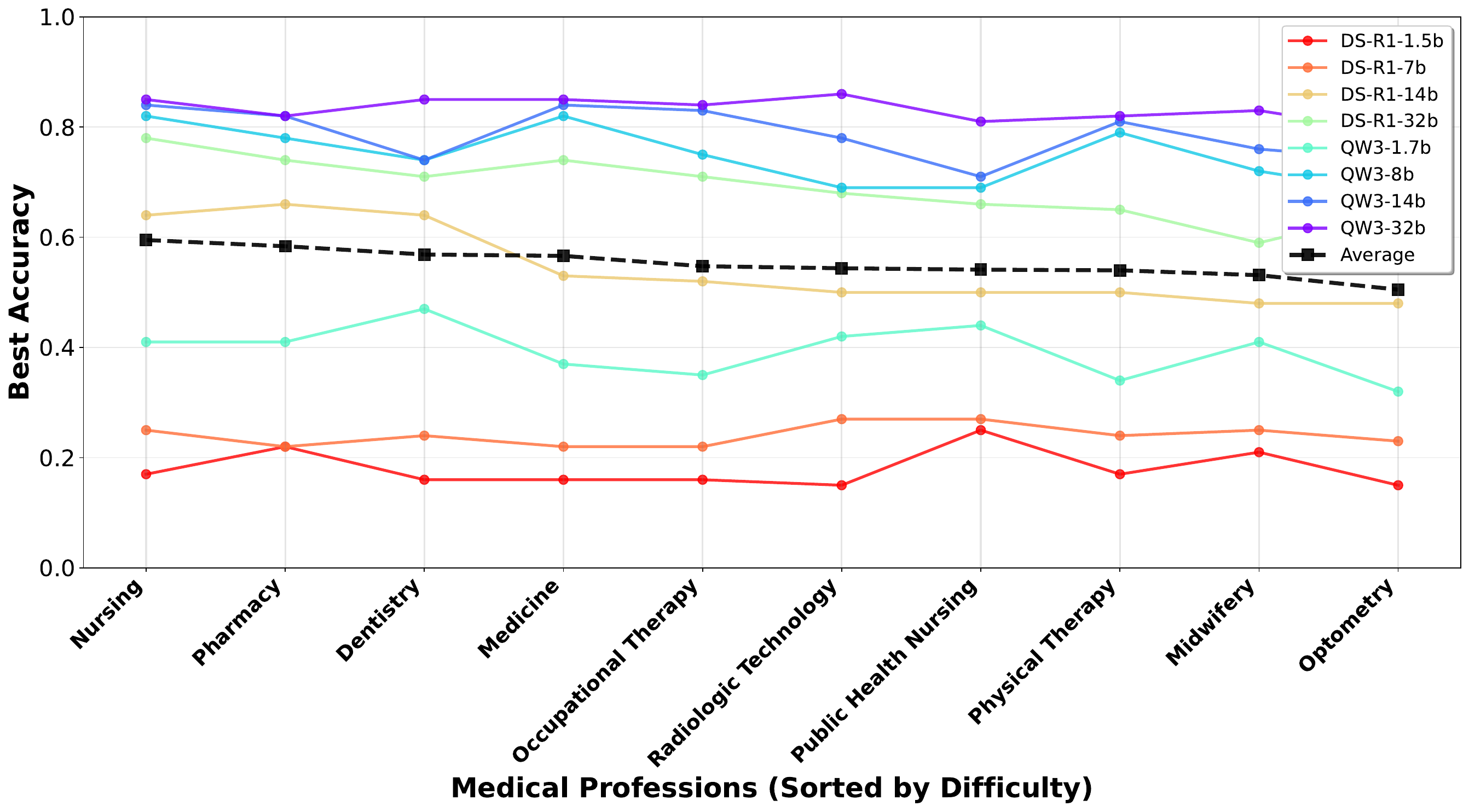}
\caption{Parallel-coordinates view of best model performance across specialties ordered by difficulty. Stable upper trajectories indicate models that maintain accuracy as questions become more demanding. The plot shows that stronger transfer configurations not only improve average accuracy but also degrade more gracefully across domains.}
\label{fig:parallel_coords}
\end{figure}

Figure~\ref{fig:parallel_coords} provides a complementary view by ordering specialties from easier to harder. The relative ranking of models remains largely stable across this difficulty gradient: larger models occupy the upper trajectories, while small models drop more steeply as tasks become harder. This suggests that model capacity contributes not only to peak accuracy but also to graceful degradation across domains.

\begin{figure}[!ht]
\centering
\includegraphics[width=\columnwidth]{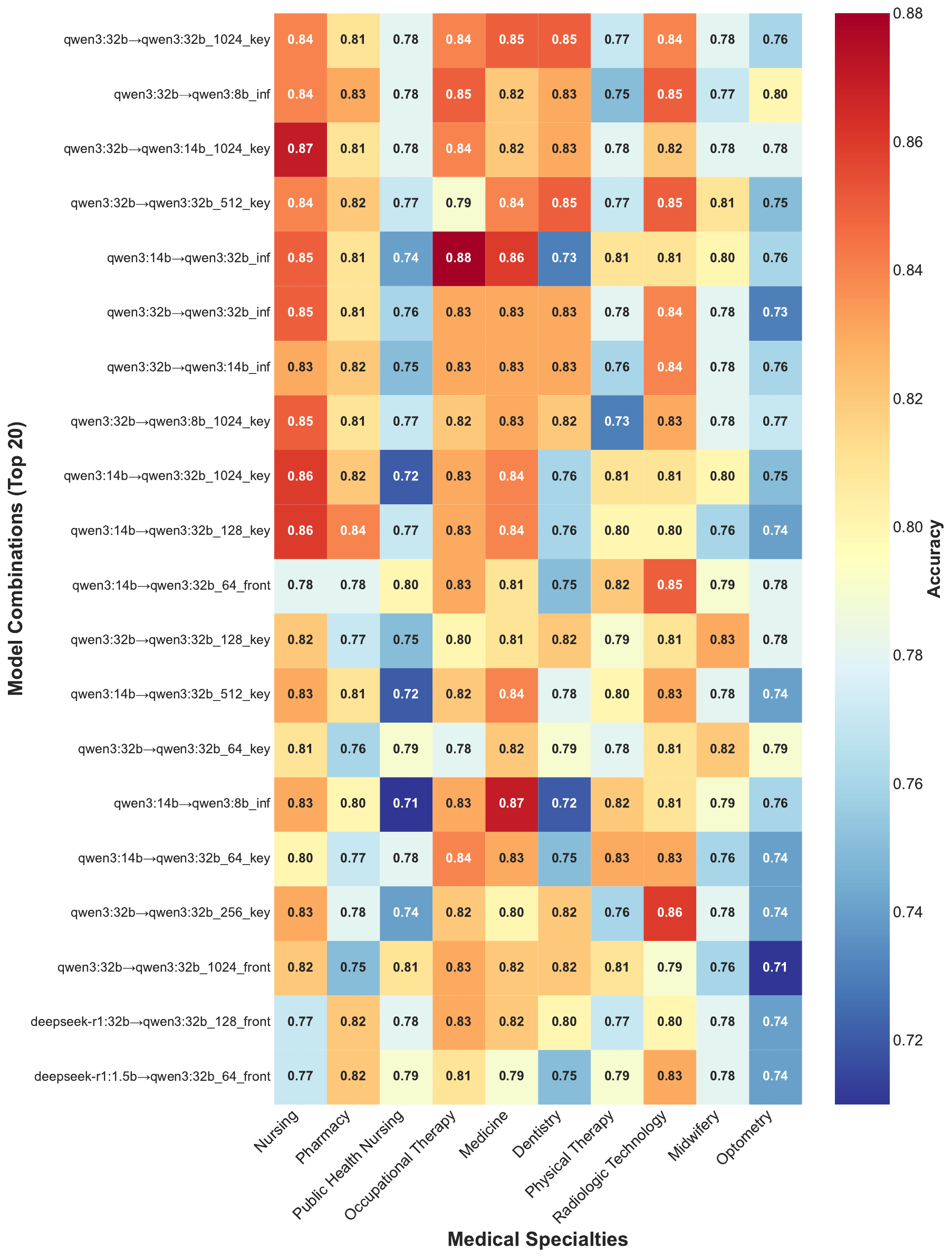}
\caption{Top-20 transfer configurations by specialty. Rows are thinking--answering model pairs with their selected budgets and strategies; columns are medical specialties ordered by difficulty. Darker cells indicate higher accuracy. The best rows are dominated by adaptive summarization and medium-to-large answerers, showing that CoT-X improves both peak performance and cross-specialty consistency.}
\label{fig:top_combinations}
\end{figure}

The top-combination heatmap in Figure~\ref{fig:top_combinations} identifies which pairings preserve this robustness after transfer. The strongest configurations usually combine a 32B or 14B thinking model with a medium-to-large answering model and adaptive summarization. The best 32B--32B configuration achieves the highest average accuracy, but several asymmetric pairings approach its performance with lower cost. These ``efficiency champions'' are important because they are the configurations a production system would prefer when accuracy and serving cost must be balanced.

\begin{figure*}[!ht]
\centering
\includegraphics[width=\textwidth]{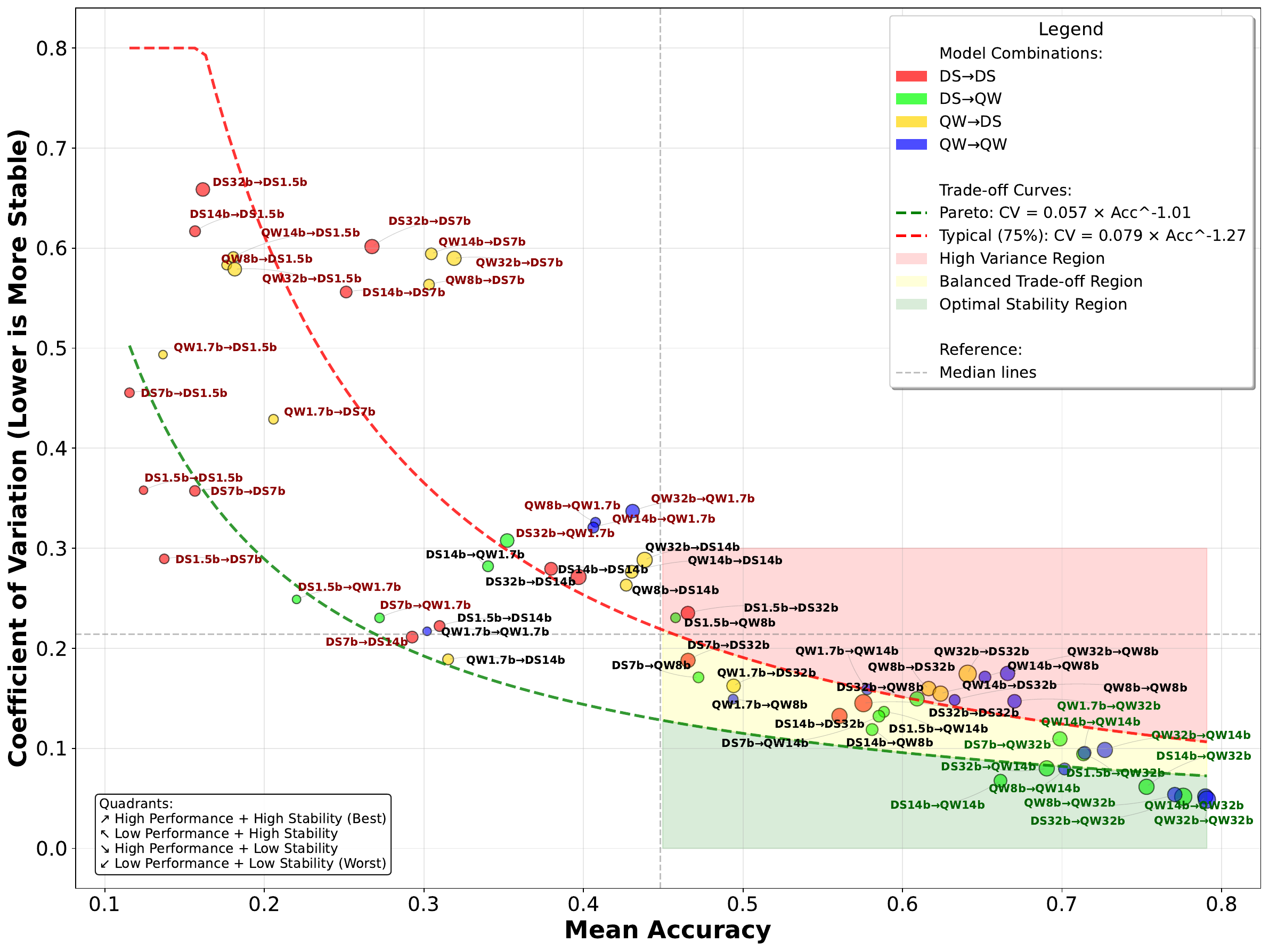}
\caption{Accuracy--robustness trade-off for all transfer configurations. Each point is a model pair; the x-axis is mean accuracy and the y-axis is coefficient of variation across specialties, where lower is more stable. The Pareto frontier identifies configurations that cannot be improved in accuracy without sacrificing robustness, providing a deployment map rather than a single leaderboard score.}
\label{fig:pareto_frontier}
\end{figure*}

Figure~\ref{fig:pareto_frontier} aggregates the domain analysis into a performance--robustness trade-off. Each point is a transfer configuration, the x-axis is mean accuracy, and the y-axis is coefficient of variation across specialties, where lower is more stable. The fitted trend follows $\text{CV}=0.42\times\text{Acc}^{-2.3}$ within the observed range, indicating that higher-accuracy configurations also tend to be more stable across specialties. The Pareto frontier highlights configurations where improving average accuracy would require accepting higher variance, or vice versa.

The frontier also clarifies the role of model family. Same-family transfers cluster closer to the frontier, while cross-family transfers are more dispersed. This does not make cross-family transfer undesirable; rather, it means cross-family deployments should be selected more carefully, preferably using the optimization procedure below instead of relying on scale alone.

\subsection{Computational Analysis}

\begin{figure}[!ht]
\centering
\includegraphics[width=\columnwidth]{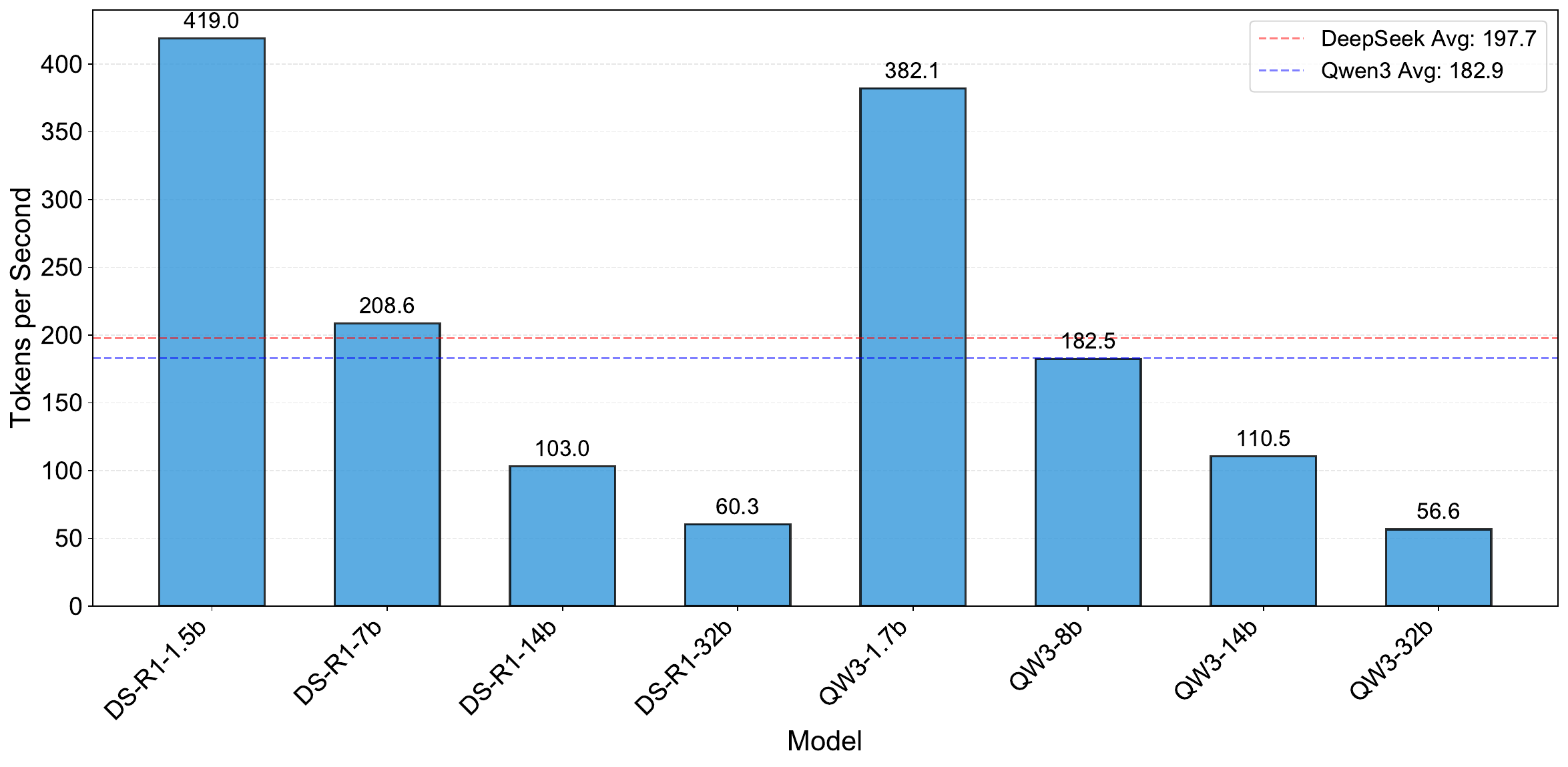}
\caption{Generation throughput measured on an 8$\times$H100 vLLM serving stack. Bars show tokens per second for each model, and dashed lines show family averages. The steep throughput drop from 1.5B--1.7B models to 32B models explains why CoT-X's asymmetric deployment pattern is attractive: expensive reasoning can be generated once and then consumed by a faster answerer.}
\label{fig:generation_speed}
\end{figure}

Figure~\ref{fig:generation_speed} shows why cross-model transfer is attractive even when large thinking models produce the best traces. Throughput drops sharply with model size: 1.5B--1.7B models exceed $380$ tokens/s, 7B--8B models reach roughly $180$--$210$ tokens/s, 14B models reach about $100$--$110$ tokens/s, and 32B models are limited to about $56$--$60$ tokens/s. Since full reasoning traces can exceed thousands of tokens, repeatedly using a 32B model for both reasoning and answering is expensive.

The deployment implication is that CoT-X should be viewed as amortization. A large model can generate a high-quality trace when needed; adaptive summarization reduces it to a few hundred tokens; and a smaller answering model consumes that compact trace at much higher throughput. Appendix Figures~\ref{fig:question_dist_full} and~\ref{fig:thinking_dist_full} further show that medical questions are typically much shorter than their generated rationales, so the compression target is the reasoning trace rather than the question itself.

\subsection{Bayesian Optimization Efficiency}

Finally, we evaluate whether good transfer configurations can be found without exhaustively testing all model pairs. The Gaussian Process Bayesian optimization module reaches $91\%$ of the exhaustive-search optimum after $8$ evaluations, $94\%$ after $10$, and $97\%$ after $15$. This corresponds to an $84\%$ reduction in evaluation cost, because only $15$ configurations are evaluated instead of all $64$.

The search trajectory is interpretable. Early evaluations sample diverse corners of the space, including small and large models as well as same-family and cross-family transfers. Later evaluations concentrate on high expected-improvement regions suggested by the surrogate. For the full dataset, this reduces the number of question--configuration evaluations from $480{,}064$ to $112{,}515$, saving more than $367{,}000$ evaluations while still identifying near-optimal deployment points.

The result is a practical workflow: use adaptive summarization as the default compression strategy, then use Bayesian optimization to choose the model pair and token budget for the target deployment constraint. This turns a large empirical grid into a small, guided search without discarding the accuracy--robustness trade-offs revealed by the full analysis.

\subsection{Cross-Language Performance Analysis}

To assess whether CoT-X depends on the source language of the exam, we extended the evaluation to Chinese and English versions of the medical QA dataset translated from the original Japanese corpus. This multilingual comparison probes a practical question: whether compressed reasoning remains useful when medical terminology, tokenization patterns, and surface syntax change across languages~\cite{qwen3,jiang2025jmedbench}.

\begin{figure}[!ht]
\centering
\includegraphics[width=\columnwidth]{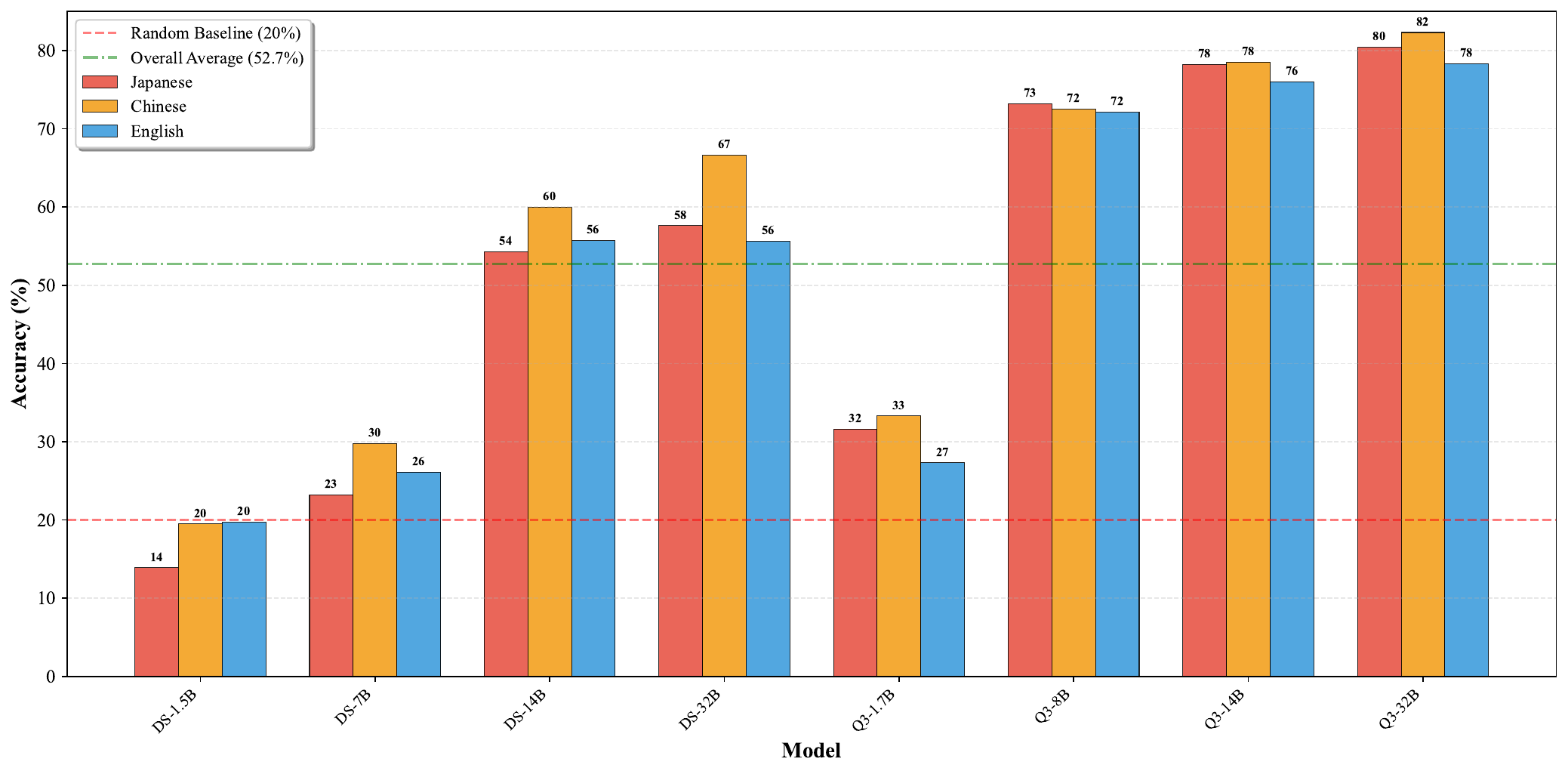}
\caption{Self-performance accuracy across Japanese, Chinese, and English versions of the medical QA dataset. Each bar uses the same model as both thinker and answerer without a compression budget, providing a language-level reference before cross-model transfer. The green dashed line marks the overall average accuracy (52.7\%), and the red line marks the five-option random baseline (20\%).}
\label{fig:language_comparison}
\end{figure}

Figure~\ref{fig:language_comparison} summarizes the performance of eight models across the three languages. Several patterns are consistent. Chinese yields the highest accuracy for most models, with an average of 55.3\%, followed by English (51.4\%) and Japanese (51.2\%). Notably, Japanese--the source language of the dataset--performs slightly worse than the translated versions, suggesting that language familiarity alone does not determine model alignment~\cite{qwen3}.

The advantage of Chinese is especially evident in smaller models. For example, DeepSeek-R1-1.5B achieves 19.5\% on Chinese compared to 13.9\% on Japanese, a 40\% relative improvement. Similarly, DeepSeek-R1-7B reaches 29.8\% on Chinese versus 23.2\% on Japanese. This suggests that Chinese text may allow more compact and semantically efficient encoding, possibly because its logographic script packs more meaning per token.

However, the language gap diminishes for larger models. Within the 32B scale, performance differences shrink to below 3\% across languages—Qwen3-32B attains 80.4\% (Japanese), 82.3\% (Chinese), and 78.3\% (English). This convergence indicates that larger models develop more language-invariant reasoning representations capable of handling medical inference across linguistic boundaries.

Across all languages, the Qwen3 family consistently outperforms DeepSeek-R1 models. The advantage is most pronounced in mid-sized configurations: Qwen3-8B exceeds DeepSeek-R1-7B by more than 40 percentage points. This can be attributed to Qwen3’s enhanced multilingual pretraining and architectural refinements, which appear to strengthen cross-lingual reasoning ability.

\subsubsection{Implications for Chain-of-Thought Transfer}

The observed language-dependent variations provide key insights into cross-lingual chain-of-thought transfer. The stable performance trends across languages confirm that the core components of our framework—adaptive summarization and model pairing—are effectively language-agnostic. This property greatly simplifies its deployment in multilingual medical contexts.

The superior performance of Chinese, particularly for smaller models, likely stems from its high information density. Its logographic system encodes medical terms more compactly, whereas alphabetic languages such as English require more tokens to convey equivalent meaning.

Most importantly for this paper, adaptive summarization consistently outperforms direct truncation across all three languages. This suggests that CoT-X is not merely exploiting Japanese-specific phrasing; it is preserving reasoning elements that remain useful after translation, which supports multilingual deployment in medical QA.

\section{Discussion}\label{d}
\subsection{Key Findings and Implications}

Our evaluation suggests that efficient CoT transfer is governed by three coupled factors: the quality of the source rationale, the compatibility between thinking and answering models, and the amount of reasoning budget available to the answerer. The main implication is that deployment should not treat CoT as an all-or-nothing capability. Long traces can be made useful to smaller models, but only when compression preserves the evidence path rather than token position.

\textbf{Adaptive summarization is most valuable under real constraints.} The largest gains over truncation occur at $64$--$256$ tokens, precisely where edge and high-throughput systems are likely to operate. This supports the central design choice of CoT-X: compression should preserve diagnostic evidence, causal links, and conclusion support, not merely the earliest tokens in a trace.

\textbf{Accuracy and robustness move together in this benchmark.} The fitted relationship $\text{CV}=0.42\times\text{Acc}^{-2.3}$ indicates that stronger configurations usually vary less across specialties. We interpret this as an empirical regularity rather than a universal law: within our model families and medical benchmark, the same configurations that capture richer reasoning also degrade more gracefully across domains. The Pareto frontier gives practitioners a concrete way to select acceptable accuracy--variance trade-offs.

\textbf{Cross-family transfer is useful but selection-sensitive.} Same-family transfer is more reliable, but cross-family transfer remains competitive after summarization. This suggests that reasoning traces contain model-agnostic information, while their surface form still reflects the source model. Adaptive summarization helps normalize this form, making hybrid deployments feasible when the best thinking model and the best answering model come from different families.

\textbf{Asymmetric pairings are often the best engineering point.} Larger models remain better thinkers and better answerers, but maximum scale on both sides is not always necessary. Pairing a large thinking model with a medium answering model can retain much of the best accuracy while substantially reducing serving cost. This is the main practical advantage of separating reasoning generation from answer generation.

\textbf{Configuration search should be guided, not exhaustive.} The Bayesian optimization results show that model-pair and budget selection can be reduced to a small number of evaluations. This matters because the transfer landscape changes as new models appear; a lightweight optimizer makes re-selection practical without rerunning the full grid.

\subsection{Practical Deployment Guidelines}

Based on these findings, we recommend choosing configurations by deployment regime rather than by model size alone.

For high-accuracy settings where compute is available, 32B thinking and 32B answering models with $512$--$1024$ summarized tokens give the strongest accuracy and low cross-specialty variance. These settings are appropriate when accuracy dominates latency and infrastructure cost.

For balanced serving, 14B--32B thinking models paired with 7B--14B answering models and a $256$-token summarized trace provide a strong accuracy--cost trade-off. This regime is the most attractive for repeated inference, because the expensive reasoning trace can be cached or amortized while the answerer remains relatively efficient.

For edge or mobile deployment, a cloud-side 7B--14B thinking model can generate reasoning for a 1.5B--1.7B local answerer. In this regime, $128$--$256$ token budgets are preferable: $64$ tokens are possible but brittle, while larger budgets may exceed local latency or memory constraints.

Regarding strategy selection, adaptive summarization should be the default below $512$ tokens, for answering models smaller than 14B, and for cross-family transfer. Direct truncation is mainly a reasonable fallback when budgets are large and the source trace is already short.

\subsection{Limitations}

Our study has several limitations. First, we use fixed temperature settings ($0.7$ for thinking models, $0.1$ for answering models, and $0.3$ for summarization); different decoding policies may change absolute accuracy. Second, our experiments use standardized CoT prompts rather than model-specific prompt tuning, so some model pairs may be under-optimized. Third, all experiments were performed on H100 GPUs using vLLM~\cite{kwon2023vllm}; relative trends should be more stable than absolute throughput, but hardware and serving stack matter for latency. Fourth, the importance weights ($\alpha_1$--$\alpha_4$) are heuristic and should be revisited for domains with different reasoning structures. Finally, the benchmark is multiple-choice medical QA, so future work should test whether the same compression strategy transfers to open-ended generation, interactive diagnosis, and tasks where the answer format is less constrained.

\subsection{Future Work}

\textbf{Lightweight compression agents.}
The current implementation uses Qwen3-32B as the summarization agent, which is effective but not the cheapest possible choice. A natural next step is to train or distill a smaller compression model specialized for reasoning traces. Such a model would not need broad open-ended generation ability; it would mainly need to identify diagnostic evidence, preserve causal links, and rewrite the selected segments compactly. This could reduce preprocessing cost and make CoT-X more attractive for high-throughput serving.

\textbf{Broader model and API coverage.}
Our controlled grid focuses on open-weight DeepSeek-R1 and Qwen3 models so that throughput and memory can be measured directly. Future work should extend the same protocol to API-served systems such as GPT-4, Claude, and Gemini, where fine-tuning is often unavailable but inference-time context control remains possible~\cite{openai2023gpt4,anthropic2024claude3,geminiteam2023gemini}. This would also test whether compressed rationales transfer between closed and open models, a common deployment pattern in real products.

\textbf{Multimodal and structured clinical reasoning.}
Medical reasoning often depends on images, tables, waveforms, longitudinal records, and structured laboratory values. Extending CoT-X beyond text would require aligning verbal rationales with multimodal evidence while preserving provenance. A useful direction is to treat radiology images, lab panels, and clinical tables as typed evidence nodes, then compress both the reasoning text and the supporting evidence references.

\textbf{Interactive and continually updated reasoning.}
In many medical and enterprise settings, answers are produced over several turns rather than from a single static question. Future versions of CoT-X should support incremental updates: when new evidence arrives, the system should revise only the affected parts of the compressed rationale instead of regenerating the whole trace. This would make the framework better suited to clinical triage, tutoring, and agent workflows where context changes over time.

\textbf{Wider benchmark validation and safety analysis.}
The performance--robustness relationship observed in this paper should be tested on additional medical and general reasoning benchmarks, including MedQA, PubMedQA, MedMCQA, MMLU, BIG-bench, and multilingual Japanese biomedical benchmarks~\cite{jin2021medqa,jin2019pubmedqa,pal2022medmcqa,hendrycks2021mmlu,srivastava2023bigbench,jiang2025jmedbench,liu2025kokushimd}. Beyond accuracy, future work should study whether compression changes calibration, uncertainty expression, or safety-critical omissions. In medical applications, a shorter rationale is useful only if it remains faithful to the evidence and does not hide clinically important uncertainty.

\section{Conclusion}\label{c}
This paper introduced CoT-X, an adaptive framework for transferring long Chain-of-Thought traces from capable thinking models to efficient answering models. Across $7,501$ medical licensing questions and $64$ DeepSeek-R1/Qwen3 transfer pairs, adaptive summarization consistently outperforms direct truncation, with the largest gains under tight $64$--$256$ token budgets. The experiments show that successful CoT transfer is shaped by both compatibility and compression: same-family transfers are most reliable, cross-family transfers remain viable after summarization, and asymmetric pairings can retain strong accuracy at lower serving cost. The Gaussian Process Bayesian optimization layer further reduces configuration search by $84\%$, making deployment-oriented selection practical. Taken together, the results suggest that long-form reasoning does not need to be regenerated by every model at inference time. With budget-aware compression and guided model selection, CoT-style reasoning can be reused in systems that must balance accuracy, latency, and cost.

\bibliographystyle{IEEEtran}
\bibliography{references}

@inproceedings{wei2022chain,
  title={Chain-of-Thought Prompting Elicits Reasoning in Large Language Models},
  author={Wei, Jason and Wang, Xuezhi and Schuurmans, Dale and Bosma, Maarten and Ichter, Brian and Xia, Fei and Chi, Ed H. and Le, Quoc V. and Zhou, Denny},
  booktitle={Advances in Neural Information Processing Systems},
  year={2022},
  url={https://arxiv.org/abs/2201.11903}
}

@inproceedings{kojima2023large,
  title={Large Language Models are Zero-Shot Reasoners},
  author={Kojima, Takeshi and Gu, Shixiang Shane and Reid, Machel and Matsuo, Yutaka and Iwasawa, Yusuke},
  booktitle={Advances in Neural Information Processing Systems},
  year={2023},
  url={https://arxiv.org/abs/2205.11916}
}

@inproceedings{cobbe2021gsm8k,
  title={Training Verifiers to Solve Math Word Problems},
  author={Cobbe, Karl and Kosaraju, Vineet and Bavarian, Mohammad and Chen, Mark and Jun, Heewoo and Kaiser, Lukasz and Plappert, Matthias and Tworek, Jerry and Hilton, Jacob and Nakano, Reiichiro and Hesse, Christopher and Schulman, John},
  booktitle={arXiv preprint arXiv:2110.14168},
  year={2021},
  url={https://arxiv.org/abs/2110.14168}
}

@article{singhal2023clinical,
  title={Large Language Models Encode Clinical Knowledge},
  author={Singhal, Karan and Azizi, Shekoofeh and Tu, Tao and Mahdavi, S. Sara and Wei, Jason and Chung, Hyung Won and Scales, Nathan and Tanwani, Ajay and Cole-Lewis, Heather and Pfohl, Stephen and others},
  journal={Nature},
  volume={620},
  pages={172--180},
  year={2023},
  url={https://www.nature.com/articles/s41586-023-06291-2}
}

@article{jin2021medqa,
  title={What Disease does this Patient Have? A Large-scale Open Domain Question Answering Dataset from Medical Exams},
  author={Jin, Di and Pan, Eileen and Oufattole, Nassim and Weng, Wei-Hung and Fang, Hanyi and Szolovits, Peter},
  journal={Applied Sciences},
  volume={11},
  number={14},
  pages={6421},
  year={2021},
  url={https://arxiv.org/abs/2009.13081}
}

@article{deepseekr1,
  title={DeepSeek-R1: Incentivizing Reasoning Capability in LLMs via Reinforcement Learning},
  author={{DeepSeek-AI} and Guo, Daya and Yang, Dejian and Zhang, Haowei and Song, Junxiao and Zhang, Ruoyu and Xu, Runxin and Zhu, Qihao and Ma, Shirong and Wang, Peiyi and others},
  journal={arXiv preprint arXiv:2501.12948},
  year={2025},
  url={https://arxiv.org/abs/2501.12948}
}

@article{qwen3,
  title={Qwen3 Technical Report},
  author={Yang, An and Li, Anfeng and Yang, Baosong and Zhang, Beichen and Hui, Binyuan and Zheng, Bo and Yu, Bowen and Gao, Chang and Huang, Chengen and Lv, Chenxu and others},
  journal={arXiv preprint arXiv:2505.09388},
  year={2025},
  url={https://arxiv.org/abs/2505.09388}
}

@article{yao2023treeofthought,
  title={Tree of Thoughts: Deliberate Problem Solving with Large Language Models},
  author={Yao, Shunyu and Yu, Dian and Zhao, Jeffrey and Shafran, Izhak and Griffiths, Thomas L. and Cao, Yuan and Narasimhan, Karthik},
  journal={arXiv preprint arXiv:2305.10601},
  year={2023},
  url={https://arxiv.org/abs/2305.10601}
}

@inproceedings{wang2023selfconsistency,
  title={Self-Consistency Improves Chain of Thought Reasoning in Language Models},
  author={Wang, Xuezhi and Wei, Jason and Schuurmans, Dale and Le, Quoc V. and Chi, Ed H. and Narang, Sharan and Chowdhery, Aakanksha and Zhou, Denny},
  booktitle={International Conference on Learning Representations},
  year={2023},
  url={https://arxiv.org/abs/2203.11171}
}

@inproceedings{zhou2023leasttomost,
  title={Least-to-Most Prompting Enables Complex Reasoning in Large Language Models},
  author={Zhou, Denny and Sch{\"a}rli, Nathanael and Hou, Le and Wei, Jason and Scales, Nathan and Wang, Xuezhi and Schuurmans, Dale and Cui, Claire and Bousquet, Olivier and Le, Quoc V. and Chi, Ed H.},
  booktitle={International Conference on Learning Representations},
  year={2023},
  url={https://arxiv.org/abs/2205.10625}
}

@inproceedings{jiang2023llmlingua,
  title={{LLMLingua}: Compressing Prompts for Accelerated Inference of Large Language Models},
  author={Jiang, Huiqiang and Wu, Qianhui and Lin, Chin-Yew and Yang, Yuqing and Qiu, Lili},
  booktitle={Proceedings of the 2023 Conference on Empirical Methods in Natural Language Processing},
  year={2023},
  url={https://aclanthology.org/2023.emnlp-main.825/}
}

@article{liu2024lostmiddle,
  title={Lost in the Middle: How Language Models Use Long Contexts},
  author={Liu, Nelson F. and Lin, Kevin and Hewitt, John and Paranjape, Ashwin and Bevilacqua, Michele and Petroni, Fabio and Liang, Percy},
  journal={Transactions of the Association for Computational Linguistics},
  volume={12},
  pages={157--173},
  year={2024},
  url={https://aclanthology.org/2024.tacl-1.9/}
}

@article{magister2023teaching,
  title={Teaching Small Language Models to Reason},
  author={Magister, Tomas and Mallinson, Jonathan and Adamek, Jakub and Malmi, Eric and Severyn, Aliaksei},
  journal={arXiv preprint arXiv:2212.08410},
  year={2023},
  url={https://arxiv.org/abs/2212.08410}
}

@inproceedings{hsieh2023distilling,
  title={Distilling Step-by-Step! Outperforming Larger Language Models with Less Training Data and Smaller Model Sizes},
  author={Hsieh, Cheng-Yu and Li, Chun-Liang and Yeh, Chih-Kuan and Nakhost, Hootan and Fujii, Yasuhisa and Ratner, Alexander and Krishna, Ranjay and Lee, Chen-Yu and Pfister, Tomas},
  booktitle={Findings of the Association for Computational Linguistics: ACL 2023},
  pages={8003--8017},
  year={2023},
  url={https://aclanthology.org/2023.findings-acl.507/}
}

@article{cheng2017survey,
  title={A Survey of Model Compression and Acceleration for Deep Neural Networks},
  author={Cheng, Yu and Wang, Duo and Zhou, Pan and Zhang, Tao},
  journal={arXiv preprint arXiv:1710.09282},
  year={2017},
  url={https://arxiv.org/abs/1710.09282}
}

@inproceedings{hinton2015distilling,
  title={Distilling the Knowledge in a Neural Network},
  author={Hinton, Geoffrey and Vinyals, Oriol and Dean, Jeff},
  booktitle={NeurIPS Deep Learning and Representation Learning Workshop},
  year={2015},
  url={https://arxiv.org/abs/1503.02531}
}

@inproceedings{touvron2021training,
  title={Training Data-Efficient Image Transformers \& Distillation through Attention},
  author={Touvron, Hugo and Cord, Matthieu and Douze, Matthijs and Massa, Francisco and Sablayrolles, Alexandre and J{\'e}gou, Herv{\'e}},
  booktitle={International Conference on Machine Learning},
  year={2021},
  url={https://proceedings.mlr.press/v139/touvron21a.html}
}

@article{nenkova2012survey,
  title={A Survey of Text Summarization Techniques},
  author={Nenkova, Ani and McKeown, Kathleen},
  journal={Foundations and Trends in Information Retrieval},
  volume={5},
  number={2--3},
  pages={103--233},
  year={2012}
}

@inproceedings{see2017get,
  title={Get to the Point: Summarization with Pointer-Generator Networks},
  author={See, Abigail and Liu, Peter J. and Manning, Christopher D.},
  booktitle={Proceedings of the 55th Annual Meeting of the Association for Computational Linguistics},
  pages={1073--1083},
  year={2017},
  url={https://aclanthology.org/P17-1099/}
}

@inproceedings{lewis2020bart,
  title={{BART}: Denoising Sequence-to-Sequence Pre-training for Natural Language Generation, Translation, and Comprehension},
  author={Lewis, Mike and Liu, Yinhan and Goyal, Naman and Ghazvininejad, Marjan and Mohamed, Abdelrahman and Levy, Omer and Stoyanov, Veselin and Zettlemoyer, Luke},
  booktitle={Proceedings of the 58th Annual Meeting of the Association for Computational Linguistics},
  pages={7871--7880},
  year={2020},
  url={https://aclanthology.org/2020.acl-main.703/}
}

@inproceedings{zhang2020pegasus,
  title={{PEGASUS}: Pre-training with Extracted Gap-sentences for Abstractive Summarization},
  author={Zhang, Jingqing and Zhao, Yao and Saleh, Mohammad and Liu, Peter J.},
  booktitle={International Conference on Machine Learning},
  year={2020},
  url={https://proceedings.mlr.press/v119/zhang20ae.html}
}

@article{zhang2024summarization,
  title={A Systematic Survey of Text Summarization: From Statistical Methods to Large Language Models},
  author={Zhang, Haopeng and Yu, Philip S. and Zhang, Jiawei},
  journal={arXiv preprint arXiv:2406.11289},
  year={2024},
  url={https://arxiv.org/abs/2406.11289}
}

@article{cheng2024compressedcot,
  title={Compressed Chain of Thought: Efficient Reasoning through Dense Representations},
  author={Cheng, Jeffrey and Van Durme, Benjamin},
  journal={arXiv preprint arXiv:2412.13171},
  year={2024},
  url={https://arxiv.org/abs/2412.13171}
}

@inproceedings{xia2025tokenskip,
  title={TokenSkip: Controllable Chain-of-Thought Compression in LLMs},
  author={Xia, Heming and Leong, Chak Tou and Wang, Wenjie and Li, Yongqi and Li, Wenjie},
  booktitle={Proceedings of the 2025 Conference on Empirical Methods in Natural Language Processing},
  pages={3351--3363},
  year={2025},
  url={https://aclanthology.org/2025.emnlp-main.165/}
}

@book{rasmussen2006gaussian,
  title={Gaussian Processes for Machine Learning},
  author={Rasmussen, Carl Edward and Williams, Christopher K. I.},
  year={2006},
  publisher={MIT Press},
  url={http://www.gaussianprocess.org/gpml/}
}

@inproceedings{snoek2012practical,
  title={Practical Bayesian Optimization of Machine Learning Algorithms},
  author={Snoek, Jasper and Larochelle, Hugo and Adams, Ryan P.},
  booktitle={Advances in Neural Information Processing Systems},
  year={2012},
  url={https://proceedings.neurips.cc/paper/2012/hash/05311655a15b75fab86956663e1819cd-Abstract.html}
}

@article{frazier2018tutorial,
  title={A Tutorial on Bayesian Optimization},
  author={Frazier, Peter I.},
  journal={arXiv preprint arXiv:1807.02811},
  year={2018},
  url={https://arxiv.org/abs/1807.02811}
}

@inproceedings{kandasamy2018neural,
  title={Neural Architecture Search with Bayesian Optimisation and Optimal Transport},
  author={Kandasamy, Kirthevasan and Neiswanger, Willie and Schneider, Jeff and Poczos, Barnabas and Xing, Eric P.},
  booktitle={Advances in Neural Information Processing Systems},
  year={2018},
  url={https://proceedings.neurips.cc/paper/2018/hash/f33ba15effa5c10e873bf3842afb46a6-Abstract.html}
}

@inproceedings{li2017hyperband,
  title={Hyperband: A Novel Bandit-Based Approach to Hyperparameter Optimization},
  author={Li, Lisha and Jamieson, Kevin and DeSalvo, Giulia and Rostamizadeh, Afshin and Talwalkar, Ameet},
  booktitle={International Conference on Learning Representations},
  year={2017},
  url={https://arxiv.org/abs/1603.06560}
}

@article{shahriari2016taking,
  title={Taking the Human Out of the Loop: A Review of Bayesian Optimization},
  author={Shahriari, Bobak and Swersky, Kevin and Wang, Ziyu and Adams, Ryan P. and de Freitas, Nando},
  journal={Proceedings of the IEEE},
  volume={104},
  number={1},
  pages={148--175},
  year={2016},
  url={https://doi.org/10.1109/JPROC.2015.2494218}
}

@inproceedings{page1999pagerank,
  title={The PageRank Citation Ranking: Bringing Order to the Web},
  author={Page, Lawrence and Brin, Sergey and Motwani, Rajeev and Winograd, Terry},
  booktitle={Proceedings of the 7th International World Wide Web Conference},
  year={1999},
  url={http://ilpubs.stanford.edu:8090/422/}
}

@inproceedings{maynez2020faithfulness,
  title={On Faithfulness and Factuality in Abstractive Summarization},
  author={Maynez, Joshua and Narayan, Shashi and Bohnet, Bernd and McDonald, Ryan},
  booktitle={Proceedings of the 58th Annual Meeting of the Association for Computational Linguistics},
  pages={1906--1919},
  year={2020},
  url={https://aclanthology.org/2020.acl-main.173/}
}

@inproceedings{medcot2024,
  title={MedCoT: Medical Chain of Thought via Hierarchical Expert},
  author={Liu, Jiaxiang and Wang, Yuan and Du, Jiawei and Zhou, Joey Tianyi and Liu, Zuozhu},
  booktitle={Proceedings of the 2024 Conference on Empirical Methods in Natural Language Processing},
  pages={17371--17389},
  year={2024},
  url={https://aclanthology.org/2024.emnlp-main.962/}
}

@article{lievin2024medicalreasoning,
  title={Can Large Language Models Reason about Medical Questions?},
  author={Li{\'e}vin, Valentin and Hother, Christoffer Egeberg and Motzfeldt, Andreas Geert and Winther, Ole},
  journal={Patterns},
  volume={5},
  number={3},
  pages={100943},
  year={2024},
  url={https://doi.org/10.1016/j.patter.2024.100943}
}

@article{kaplan2020scaling,
  title={Scaling Laws for Neural Language Models},
  author={Kaplan, Jared and McCandlish, Sam and Henighan, Tom and Brown, Tom B. and Chess, Benjamin and Child, Rewon and Gray, Scott and Radford, Alec and Wu, Jeffrey and Amodei, Dario},
  journal={arXiv preprint arXiv:2001.08361},
  year={2020},
  url={https://arxiv.org/abs/2001.08361}
}

@article{hernandez2021scaling,
  title={Scaling Laws for Transfer},
  author={Hernandez, Danny and Kaplan, Jared and Henighan, Tom and McCandlish, Sam},
  journal={arXiv preprint arXiv:2102.01293},
  year={2021},
  url={https://arxiv.org/abs/2102.01293}
}

@inproceedings{kwon2023vllm,
  title={Efficient Memory Management for Large Language Model Serving with PagedAttention},
  author={Kwon, Woosuk and Li, Zhuohan and Zhuang, Siyuan and Sheng, Ying and Zheng, Lianmin and Yu, Cody Hao and Gonzalez, Joseph E. and Zhang, Hao and Stoica, Ion},
  booktitle={Proceedings of the 29th ACM Symposium on Operating Systems Principles},
  pages={611--626},
  year={2023},
  url={https://doi.org/10.1145/3600006.3613165}
}

@book{cohen1988statistical,
  title={Statistical Power Analysis for the Behavioral Sciences},
  author={Cohen, Jacob},
  year={1988},
  publisher={Routledge}
}

@book{efron1994bootstrap,
  title={An Introduction to the Bootstrap},
  author={Efron, Bradley and Tibshirani, Robert J.},
  year={1994},
  publisher={Chapman \& Hall}
}

@book{who2019icd11,
  title={International Classification of Diseases 11th Revision (ICD-11)},
  author={{World Health Organization}},
  year={2019},
  publisher={World Health Organization},
  url={https://icd.who.int/}
}

@article{donnelly2006snomed,
  title={{SNOMED CT}: The Advanced Terminology and Coding System for eHealth},
  author={Donnelly, Kelvin},
  journal={Studies in Health Technology and Informatics},
  volume={121},
  pages={279--290},
  year={2006}
}

@inproceedings{zhang2020bertscore,
  title={{BERTScore}: Evaluating Text Generation with {BERT}},
  author={Zhang, Tianyi and Kishore, Varsha and Wu, Felix and Weinberger, Kilian Q. and Artzi, Yoav},
  booktitle={International Conference on Learning Representations},
  year={2020},
  url={https://openreview.net/forum?id=SkeHuCVFDr}
}

@inproceedings{fu2024gptscore,
  title={{GPTScore}: Evaluate as You Desire},
  author={Fu, Jinlan and Ng, See-Kiong and Jiang, Zhengbao and Liu, Pengfei},
  booktitle={Proceedings of the 2024 Conference of the North American Chapter of the Association for Computational Linguistics: Human Language Technologies},
  pages={6556--6576},
  year={2024},
  url={https://aclanthology.org/2024.naacl-long.365/}
}

@inproceedings{deb2002nsga2,
  title={A Fast and Elitist Multiobjective Genetic Algorithm: NSGA-II},
  author={Deb, Kalyanmoy and Pratap, Amrit and Agarwal, Sameer and Meyarivan, T.},
  booktitle={IEEE Transactions on Evolutionary Computation},
  volume={6},
  number={2},
  pages={182--197},
  year={2002},
  url={https://doi.org/10.1109/4235.996017}
}

@inproceedings{brown2020gpt3,
  title={Language Models are Few-Shot Learners},
  author={Brown, Tom B. and Mann, Benjamin and Ryder, Nick and Subbiah, Melanie and Kaplan, Jared and Dhariwal, Prafulla and Neelakantan, Arvind and Shyam, Pranav and Sastry, Girish and Askell, Amanda and others},
  booktitle={Advances in Neural Information Processing Systems},
  year={2020},
  url={https://proceedings.neurips.cc/paper/2020/hash/1457c0d6bfcb4967418bfb8ac142f64a-Abstract.html}
}

@article{openai2023gpt4,
  title={{GPT-4} Technical Report},
  author={{OpenAI} and Achiam, Josh and Adler, Steven and Agarwal, Sandhini and Ahmad, Lama and Akkaya, Ilge and Aleman, Florencia Leoni and Almeida, Diogo and Altenschmidt, Janko and Altman, Sam and others},
  journal={arXiv preprint arXiv:2303.08774},
  year={2023},
  url={https://arxiv.org/abs/2303.08774}
}

@misc{anthropic2024claude3,
  title={The {Claude} 3 Model Family: Opus, Sonnet, Haiku},
  author={{Anthropic}},
  year={2024},
  howpublished={Model card},
  url={https://www-cdn.anthropic.com/de8ba9b01c9ab7cbabf5c33b80b7bbc618857627/Model_Card_Claude_3.pdf}
}

@article{geminiteam2023gemini,
  title={Gemini: A Family of Highly Capable Multimodal Models},
  author={{Gemini Team} and Anil, Rohan and Borgeaud, Sebastian and Alayrac, Jean-Baptiste and Yu, Jiahui and Soricut, Radu and Schalkwyk, Johan and Dai, Andrew M. and Hauth, Anja and Millican, Katie and others},
  journal={arXiv preprint arXiv:2312.11805},
  year={2023},
  url={https://arxiv.org/abs/2312.11805}
}

@inproceedings{hendrycks2021mmlu,
  title={Measuring Massive Multitask Language Understanding},
  author={Hendrycks, Dan and Burns, Collin and Basart, Steven and Zou, Andy and Mazeika, Mantas and Song, Dawn and Steinhardt, Jacob},
  booktitle={International Conference on Learning Representations},
  year={2021},
  url={https://openreview.net/forum?id=d7KBjmI3GmQ}
}

@article{srivastava2023bigbench,
  title={Beyond the Imitation Game: Quantifying and Extrapolating the Capabilities of Language Models},
  author={Srivastava, Aarohi and Rastogi, Abhinav and Rao, Abhishek and Shoeb, Abu Awal Md and Abid, Abubakar and Fisch, Adam and Brown, Adam R. and Santoro, Adam and Gupta, Aditya and Garriga-Alonso, Adri{\`a} and others},
  journal={Transactions on Machine Learning Research},
  year={2023},
  url={https://arxiv.org/abs/2206.04615}
}

@article{liang2023helm,
  title={Holistic Evaluation of Language Models},
  author={Liang, Percy and Bommasani, Rishi and Lee, Tony and Tsipras, Dimitris and Soylu, Dilara and Yasunaga, Michihiro and Zhang, Yian and Narayanan, Deepak and Wu, Yuhuai and Kumar, Ananya and others},
  journal={Transactions on Machine Learning Research},
  year={2023},
  url={https://arxiv.org/abs/2211.09110}
}

@inproceedings{jin2019pubmedqa,
  title={{PubMedQA}: A Dataset for Biomedical Research Question Answering},
  author={Jin, Qiao and Dhingra, Bhuwan and Liu, Zhengping and Cohen, William W. and Lu, Xinghua},
  booktitle={Proceedings of the 2019 Conference on Empirical Methods in Natural Language Processing},
  pages={2567--2577},
  year={2019},
  url={https://aclanthology.org/D19-1259/}
}

@inproceedings{pal2022medmcqa,
  title={{MedMCQA}: A Large-scale Multi-Subject Multi-Choice Dataset for Medical Domain Question Answering},
  author={Pal, Ankit and Umapathi, Logesh Kumar and Sankarasubbu, Malaikannan},
  booktitle={Proceedings of the Conference on Health, Inference, and Learning},
  pages={248--260},
  year={2022},
  url={https://proceedings.mlr.press/v174/pal22a.html}
}

@article{nori2023capabilities,
  title={Capabilities of {GPT-4} on Medical Challenge Problems},
  author={Nori, Harsha and King, Nicholas and McKinney, Scott Mayer and Carignan, Dean and Horvitz, Eric},
  journal={arXiv preprint arXiv:2303.13375},
  year={2023},
  url={https://arxiv.org/abs/2303.13375}
}

@article{kasai2023japanese,
  title={Evaluating {GPT-4} and {ChatGPT} on Japanese Medical Licensing Examinations},
  author={Kasai, Jungo and Kasai, Yuhei and Sakaguchi, Keisuke and Yamada, Yutaro and Radev, Dragomir},
  journal={arXiv preprint arXiv:2303.18027},
  year={2023},
  url={https://arxiv.org/abs/2303.18027}
}

@inproceedings{jiang2025jmedbench,
  title={{JMedBench}: A Benchmark for Evaluating Japanese Biomedical Large Language Models},
  author={Jiang, Junfeng and Huang, Jiahao and Aizawa, Akiko},
  booktitle={Proceedings of the 31st International Conference on Computational Linguistics},
  pages={5918--5935},
  year={2025},
  url={https://aclanthology.org/2025.coling-main.395/}
}

@article{liu2025kokushimd,
  title={{KokushiMD-10}: Benchmark for Evaluating Large Language Models on Ten Japanese National Healthcare Licensing Examinations},
  author={Liu, Junyu and Yan, Kaiqi and Wang, Tianyang and Niu, Qian and Nagai-Tanima, Momoko and Aoyama, Tomoki},
  journal={arXiv preprint arXiv:2506.11114},
  year={2025},
  url={https://arxiv.org/abs/2506.11114}
}

\appendix
\clearpage
\section{Supplementary Figures}

\subsection{Token Distribution Analysis}

\begin{figure*}[!ht]
\centering
\includegraphics[width=\textwidth]{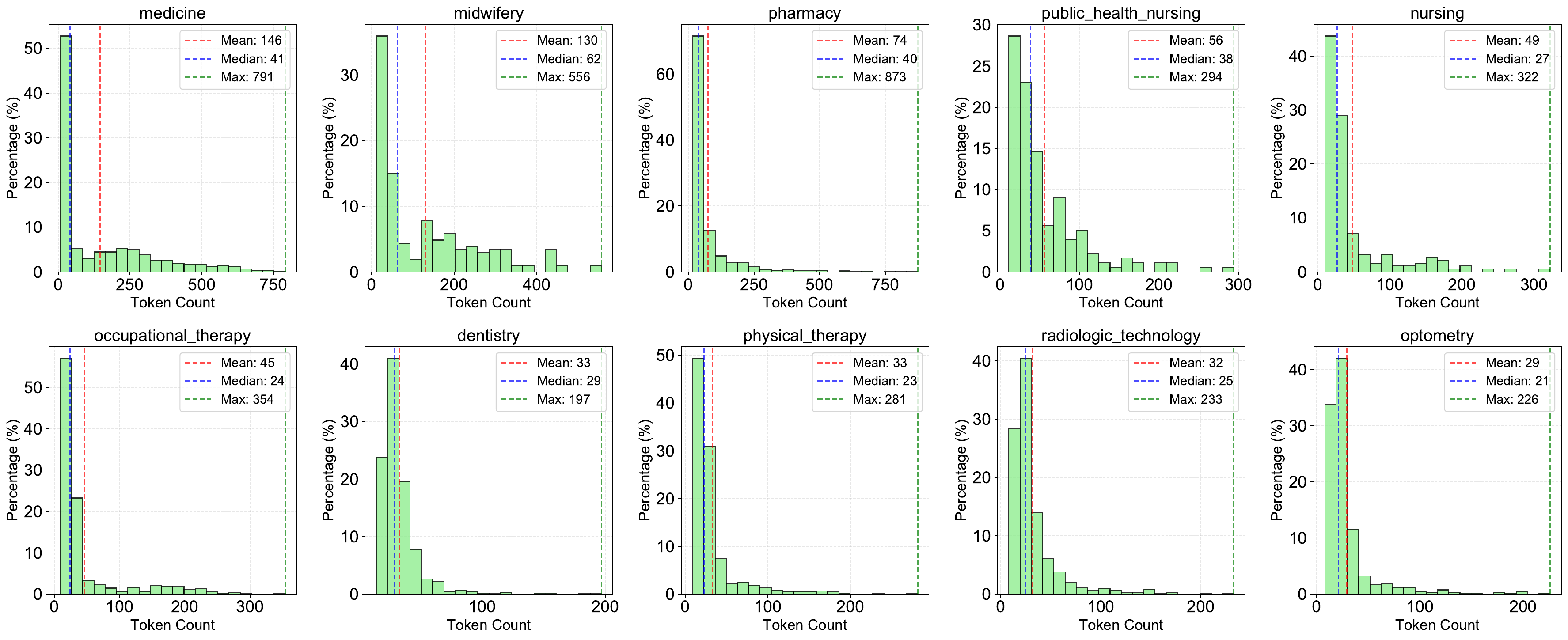}
\caption{Question-token distribution across the ten medical specialties. Each histogram reports percentage frequency and marks the mean, median, and maximum length for that specialty. Medicine and Pharmacy contain longer clinical narratives, while Optometry and Physical Therapy are shorter on average; this motivates specialty-level robustness analysis rather than relying only on aggregate accuracy.}
\label{fig:question_dist_full}
\end{figure*}

\begin{figure*}[!ht]
\centering
\includegraphics[width=\textwidth]{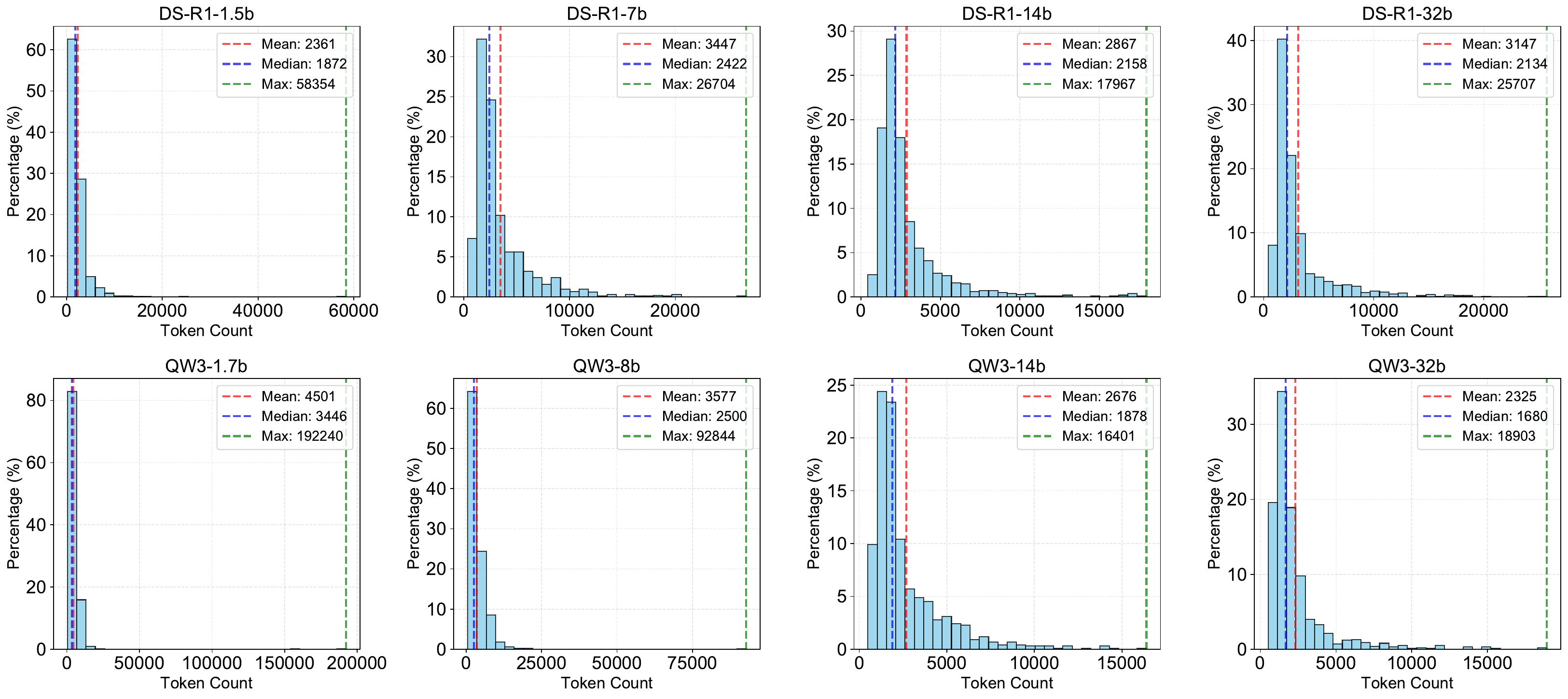}
\caption{Token-length distributions of model-generated reasoning traces. DeepSeek-R1 models, especially larger variants, produce longer and more variable chains, while Qwen3 models tend to be more compact. The gap between these raw trace lengths and the $64$--$1024$ token deployment budgets explains why selecting important reasoning content is more effective than preserving a prefix.}
\label{fig:thinking_dist_full}
\end{figure*}

Understanding the distribution of tokens in both questions and reasoning chains is essential for optimizing compression strategies and determining appropriate token budgets. Figures~\ref{fig:question_dist_full} and~\ref{fig:thinking_dist_full} present detailed histograms illustrating these distributions across different medical specialties and model types.

The question token distribution (Figure~\ref{fig:question_dist_full}) shows substantial variation among medical specialties. Questions in Medicine and Pharmacy are the longest on average, often exceeding 500 tokens due to detailed patient histories, laboratory data, and complex clinical narratives. In contrast, Optometry and Physical Therapy questions are generally concise (100–200 tokens), focusing on specific diagnostic or therapeutic scenarios.

The reasoning chain distribution (Figure~\ref{fig:thinking_dist_full}) exhibits even greater variability, which has critical implications for compression. DeepSeek-R1 models, particularly the 32B variant, produce the most extensive reasoning chains, with median lengths around 3000 tokens and maximum lengths exceeding 20000 tokens. This verbosity reflects the model’s tendency to explore multiple reasoning paths and offer detailed justifications. Qwen3-32B models generate slightly more compact reasoning, achieving approximately 20\% shorter median lengths and higher information density.

The significant gap between typical reasoning chain lengths (500–5000 tokens) and practical deployment budgets (256–512 tokens) highlights the necessity of effective compression. Even with a generous budget of 1024 tokens, about 50\% of long reasoning chains require compression, whereas at 256 tokens nearly all must be reduced by more than 50\%. This motivates our adaptive summarization approach, which substantially outperforms truncation by selectively retaining critical reasoning components.

\subsection{Implementation Details}

For reproducibility, additional implementation details are provided here. All models were loaded in half precision (float16) to optimize memory usage while preserving numerical stability. The vLLM framework~\cite{kwon2023vllm} was configured with a block size of 16 tokens and employed PagedAttention~\cite{kwon2023vllm} for efficient KV-cache management~\cite{kwon2023vllm}. Request batching was dynamically adjusted according to available GPU memory, with batch sizes ranging from 4 for 32B models to 32 for 1.5B models.

The adaptive summarization agent (Qwen3-32B) ran on a dedicated GPU for parallel compression processing. The compression instructions emphasize preservation of medical terminology, logical flow, and information directly relevant to diagnostic conclusions.

\subsection{Intelligent Summarization Agent Design}

\subsubsection{Core Innovation: Hierarchical Key Information Extraction}

The intelligent summarization agent introduces a hierarchical framework for chain-of-thought compression. It performs multi-stage semantic filtering and importance evaluation to extract and preserve essential reasoning steps under strict token constraints.

\subsubsection{Technical Framework}

\textbf{Semantic Segmentation and Importance Scoring.}  
The Qwen3-32B model segments the reasoning chain into semantically coherent units and evaluates each segment along four key dimensions defined in Equation~\eqref{eq:2}: reasoning depth $D(s_i)$, knowledge density $K(s_i)$, logical connectivity $L(s_i)$, and conclusion relevance $C(s_i)$. These scores identify portions containing core reasoning and domain-specific knowledge.

\textbf{Dynamic Compression Strategy.}  
After segmentation, the system applies several compression mechanisms, including dependency graph–based critical path extraction, redundancy elimination, entropy-based content selection, and adaptive granularity control that aligns retained content with token budget limits.

\textbf{Coherence Reconstruction.}  
Finally, the system reconstructs the compressed reasoning to ensure logical continuity and linguistic smoothness. It repairs causal dependencies, generates concise transitions, and normalizes output format to maximize token efficiency while maintaining readability.

\subsubsection{Algorithmic Innovations}

\textbf{Importance Propagation Algorithm.}  
We implement a PageRank-inspired propagation algorithm (Equation~\eqref{eq:3}) to compute global importance. Each node is initialized with a semantic importance score, and scores are iteratively propagated through the dependency graph until convergence. A normalization step ensures consistent scaling across all nodes.

\textbf{Adaptive Threshold Mechanism.}  
Retention thresholds are automatically adjusted to fit available token budgets. For instance, at 64 tokens, only the top 5\% of diagnostic reasoning is retained, while at 256 tokens roughly 30\% of core content remains. Budgets of 512 and 1024 tokens preserve 50\% and 75\% of detailed reasoning, respectively. This mechanism effectively eliminates redundant and overturned reasoning while maintaining completeness.

\textbf{Semantic Integrity Guarantee.}  
Semantic completeness is ensured through several safeguards: preservation of key medical entities, maintenance of causal “because... therefore...” structures, numerical precision retention, and consistent medical terminology usage.

\subsubsection{Comparison with Direct Truncation}

Table~\ref{tab:compression_comparison} highlights the performance benefits of our intelligent summarization approach over simple truncation.

\begin{table*}[!ht]
\centering
\caption{Qualitative comparison between direct truncation and adaptive summarization. Direct truncation preserves the beginning of a trace regardless of diagnostic importance, often interrupting the evidence path. CoT-X instead selects and reconstructs high-value reasoning segments, improving information retention, logical completeness, and medical concept preservation under the same token budget.}
\label{tab:compression_comparison}
\begin{tabular}{lcc}
\toprule
\hline
\textbf{Metric} & \textbf{Direct Truncation} & \textbf{Intelligent Summarization} \\
\midrule
Information Retention & 20–30\% & 75–85\% \\
Logical Completeness & Often Interrupted & Fully Maintained \\
Key Information Location & Random / Front-biased & Precisely Targeted \\
Adaptability & None & Content-adaptive \\
Reasoning Chain Integrity & Fragmented & Complete \\
Medical Concept Preservation & Often Lost & Fully Preserved \\
\hline
\bottomrule
\end{tabular}
\end{table*}

\subsubsection{Implementation Details and Configuration}

\textbf{Model Configuration.}  
The system uses Qwen3-32B (non-CoT variant) optimized for summarization. Inference operates in zero-shot mode with few-shot exemplars for consistency. The temperature is fixed at 0.3, and Top-p is set to 0.95, balancing diversity and accuracy. The output length is dynamically adjusted to meet token constraints.

\textbf{Compression Instruction Design.}  
The summarization agent receives structured compression instructions with explicit quality criteria and output-format constraints. The design emphasizes (1) retention of critical concepts and evidence, (2) preservation of logical flow, (3) focus on reasoning relevant to conclusions, and (4) strict adherence to standardized medical terminology.

\subsubsection{Evaluation Metrics}

Our evaluation framework employs four complementary metrics aligned with the scoring functions in our methodology~\cite{zhang2020bertscore,fu2024gptscore}.

\textbf{Information Retention Score (IRS)} measures how much critical information is preserved relative to expert-annotated references, weighted by information type.

\textbf{Logical Coherence Score (LCS)} assesses causal consistency and step continuity using automated logic detection combined with human evaluation.

\textbf{Compression Efficiency (CE)} quantifies information density per token, evaluating how effectively information is represented within budget limits.

\textbf{Medical Accuracy (MA)} validates domain-specific correctness through cross-referencing with professional medical knowledge bases.

\subsubsection{Medical Domain Optimizations}

\textbf{Medical Terminology Processing.}  
A comprehensive terminology dictionary ensures complete preservation of diagnostic terms, drug names, and abbreviations, maintaining full name–abbreviation consistency.

\textbf{Diagnostic Reasoning Chain Protection.}  
The agent prioritizes retention of the full symptom $\rightarrow$ sign $\rightarrow$examination$\rightarrow$diagnosis chain, including differential diagnoses and treatment rationale.

\textbf{Numerical Information Processing.}  
All laboratory values, dosages, and physiological measures are preserved with unit consistency and precision, as these are vital to medical reasoning.

\subsubsection{Key Technical Contributions}

The proposed intelligent summarization agent contributes several advances~\cite{cheng2024compressedcot}.  
First, it establishes the first semantic-importance-based framework for chain-of-thought compression, surpassing truncation through structured information filtering.  
Second, it utilizes non-CoT large models as compression agents, leveraging semantic comprehension without CoT overhead.  
Third, its multi-stage information extraction architecture offers modular extensibility.  
Fourth, adaptive token allocation dynamically optimizes budget utilization.  
Finally, the system achieves high compression efficiency while maintaining reasoning integrity, providing a viable solution for resource-constrained large model deployment in professional domains such as medicine~\cite{xia2025tokenskip}.

This intelligent summarization agent thus represents a practical and effective step toward deploying large reasoning models under strict computational constraints, particularly in medical and other knowledge-intensive applications.

\subsection{Complete Results for Chinese and English Datasets}

This section reports the complete experimental results for the Chinese and English datasets, following the same analysis framework used for the original Japanese corpus. These results further evaluate the generalizability of our chain-of-thought transfer framework across different linguistic settings.

\subsubsection{Chinese Dataset Results}

The Chinese version of the medical QA dataset retains the same structure and difficulty distribution as the Japanese source, containing 7,501 questions across ten medical specialties. The translation was carried out by professional medical translators and subsequently reviewed by domain experts to ensure terminological precision and clinical fidelity.

\begin{figure*}[!ht]
\centering
\includegraphics[width=0.6\textwidth]{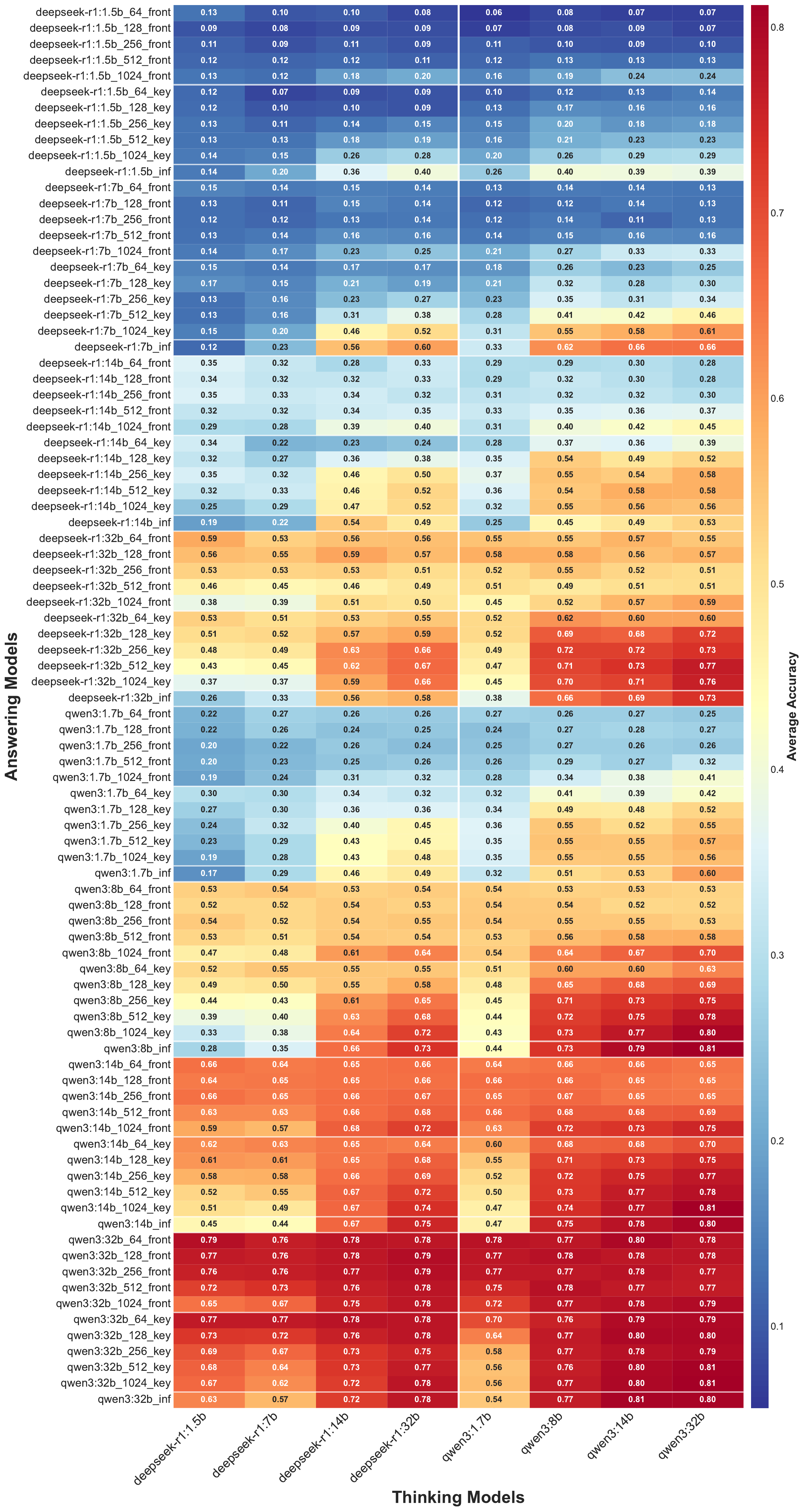}
\caption{Complete transfer matrix for the original Japanese dataset. Cells summarize accuracy across thinking--answering model combinations, token budgets, and compression strategies. Darker red indicates higher accuracy, and the stronger diagonal blocks show the advantage of same-family transfer.}
\label{fig:performance_matrix_full}
\end{figure*}

\begin{figure}[!ht]
\centering
\includegraphics[width=\columnwidth]{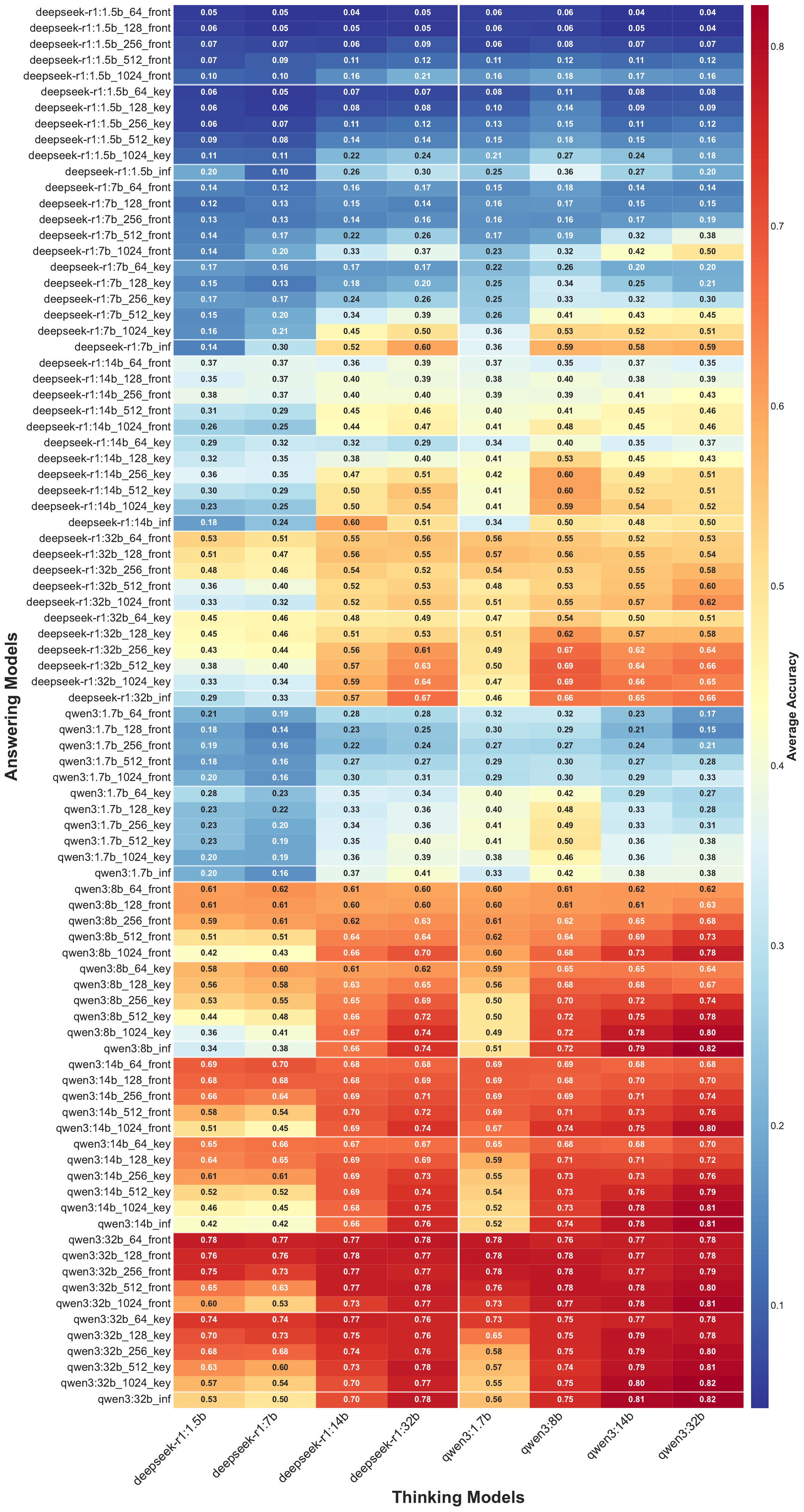}
\caption{Chinese dataset transfer matrix across thinking--answering model combinations and token budgets. Diagonal blocks remain strong, indicating that model-family compatibility persists after translation while adaptive summarization preserves useful reasoning across languages.}
\label{fig:cn_performance_matrix}
\end{figure}

Figure~\ref{fig:cn_performance_matrix} presents the performance evaluation on the Chinese dataset. The observed structure and accuracy trends closely mirror those in the Japanese and English experiments. The prominent diagonal patterns highlight the role of architectural compatibility in reasoning transfer effectiveness.

\begin{figure}[!ht]
\centering
\includegraphics[width=\columnwidth]{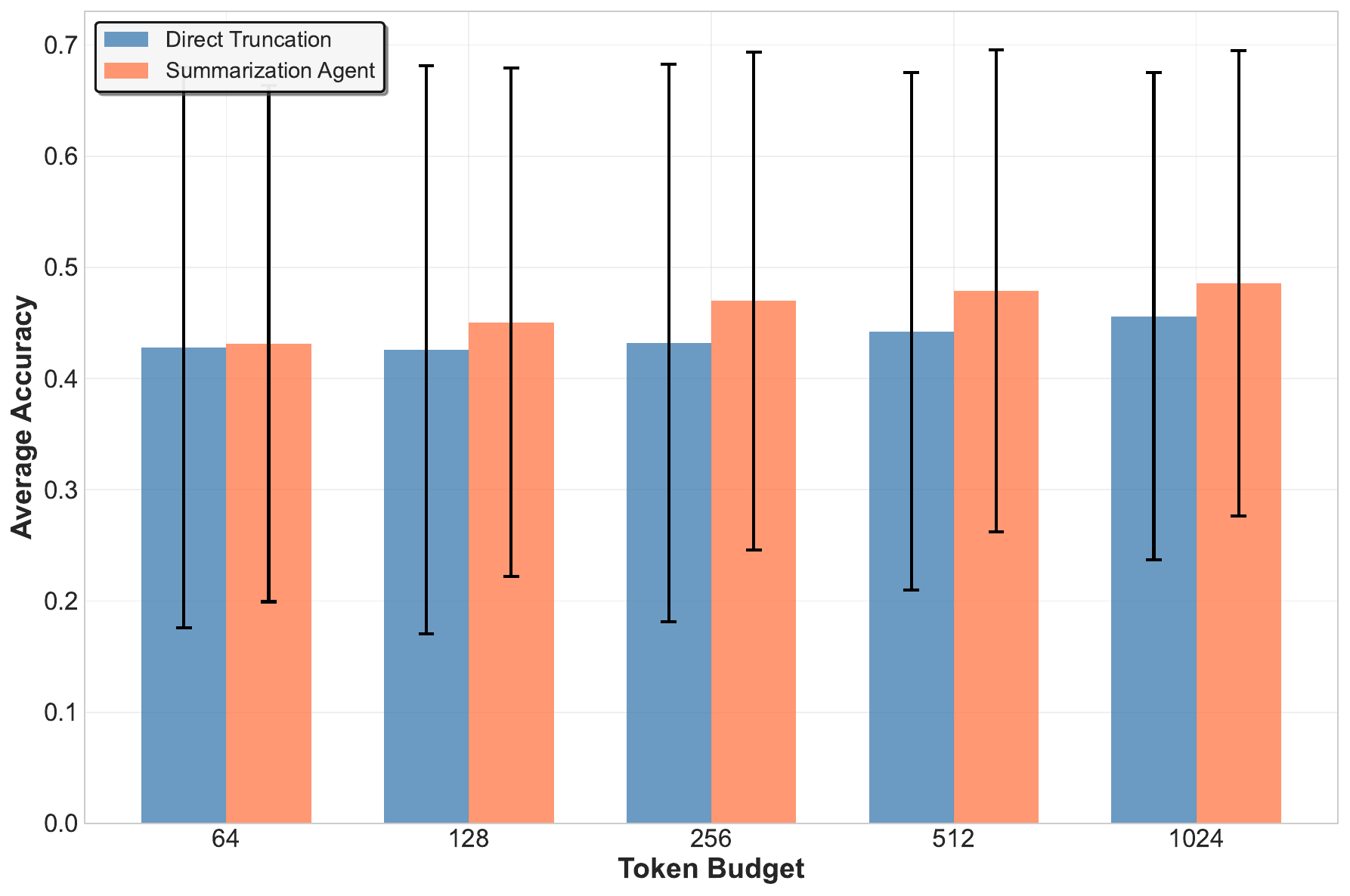}
\caption{Chinese dataset strategy comparison across token budgets. Accuracy improves as budgets increase, but the main advantage of adaptive summarization appears in the low- and mid-budget regimes where truncation loses important reasoning steps.}
\label{fig:cn_strategy_comparison}
\end{figure}

As shown in Figure~\ref{fig:cn_strategy_comparison}, model performance improves progressively as token budgets increase from 64 to 1024 tokens. The results suggest that larger token capacities enable more complete reasoning reconstruction and information retention.

\begin{figure}[!ht]
\centering
\includegraphics[width=\columnwidth]{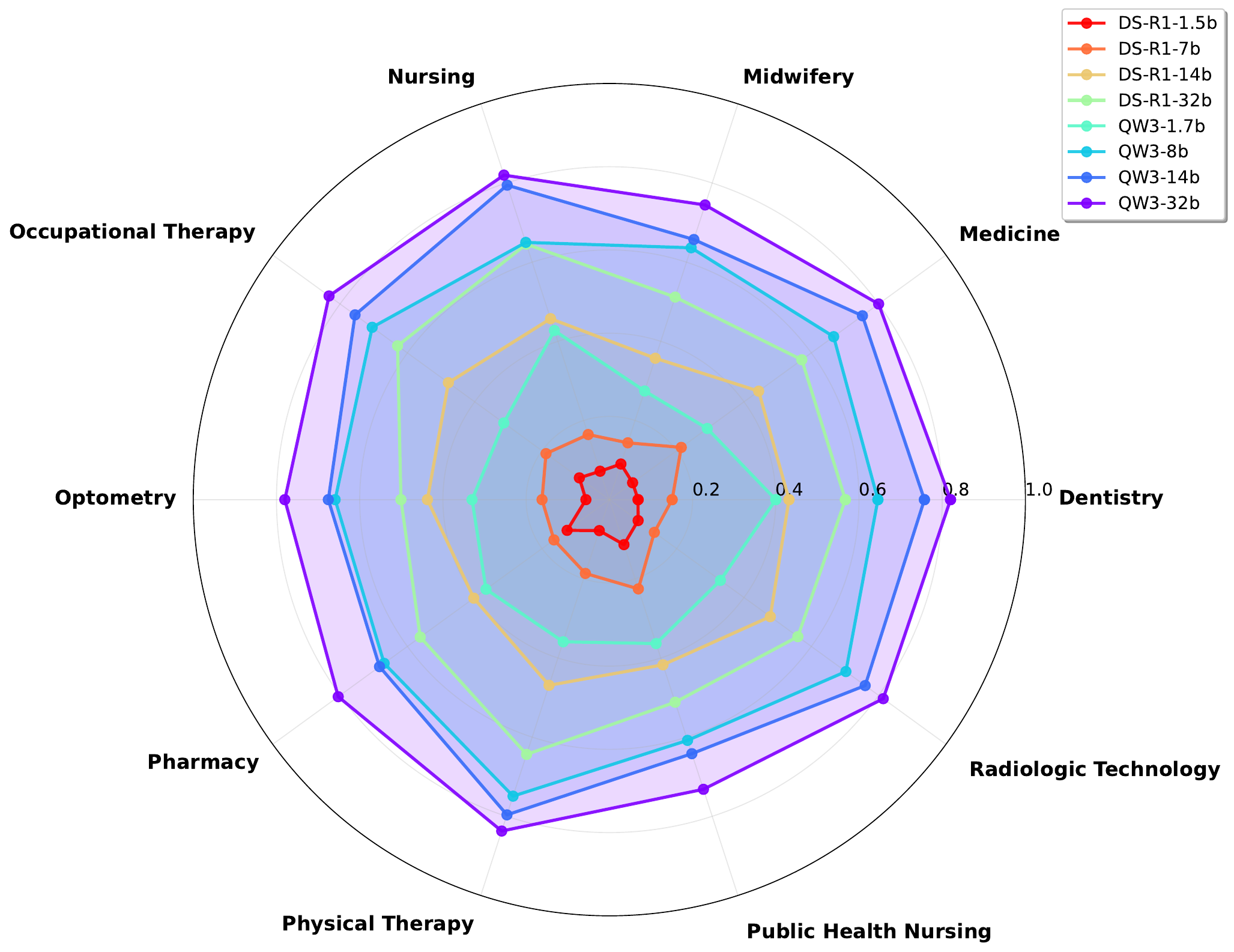}
\caption{Chinese dataset specialty profile for all evaluated models. The radar plot shows which domains are stable and which domains remain challenging after translation, supporting the cross-specialty robustness analysis.}
\label{fig:cn_radar_chart}
\end{figure}

The cross-specialty analysis (Figure~\ref{fig:cn_radar_chart}) reveals domain-level performance differences. Nursing and Physical Therapy achieve relatively higher accuracy, while Midwifery performs lower. These variations likely reflect a combination of question complexity and domain-specific knowledge representation.

\subsubsection{English Dataset Results}

The English translation underwent the same quality control process, emphasizing consistency with international medical terminologies (ICD-11~\cite{who2019icd11}, SNOMED CT~\cite{donnelly2006snomed}). The dataset preserves the original question–answer structure while adapting cultural and clinical nuances to English-language practice.

\begin{figure}[!ht]
\centering
\includegraphics[width=\columnwidth]{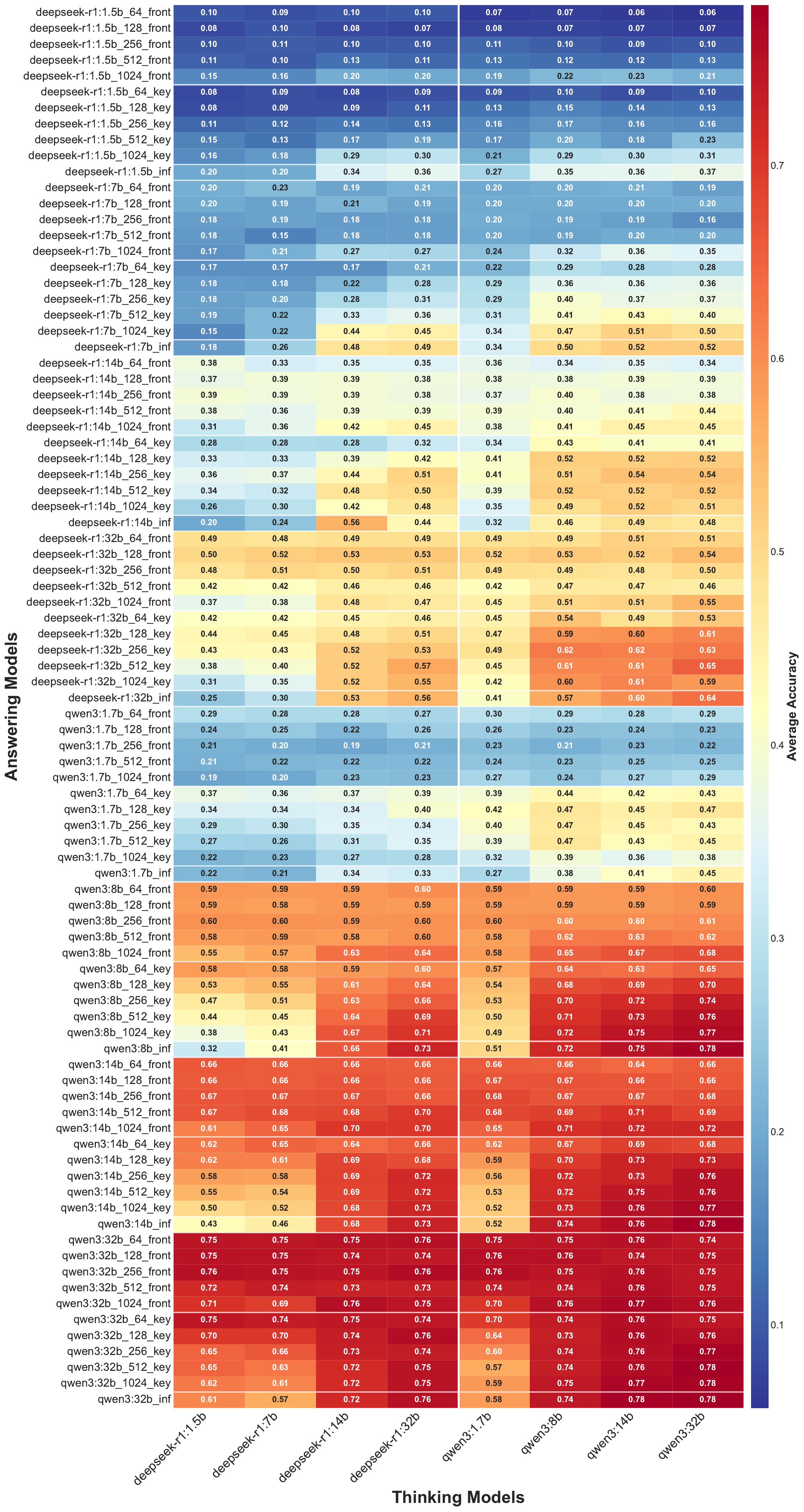}
\caption{English dataset transfer matrix across thinking--answering model pairs. The same-family blocks and asymmetric cross-family behavior remain visible, showing that the transfer structure is not specific to the Japanese source text.}
\label{fig:en_performance_matrix}
\end{figure}

Figure~\ref{fig:en_performance_matrix} shows the English dataset results. The structural similarities across languages indicate that reasoning transfer behavior remains stable. However, cross-family transfer effectiveness varies depending on specific model pairings.

\begin{figure}[!ht]
\centering
\includegraphics[width=\columnwidth]{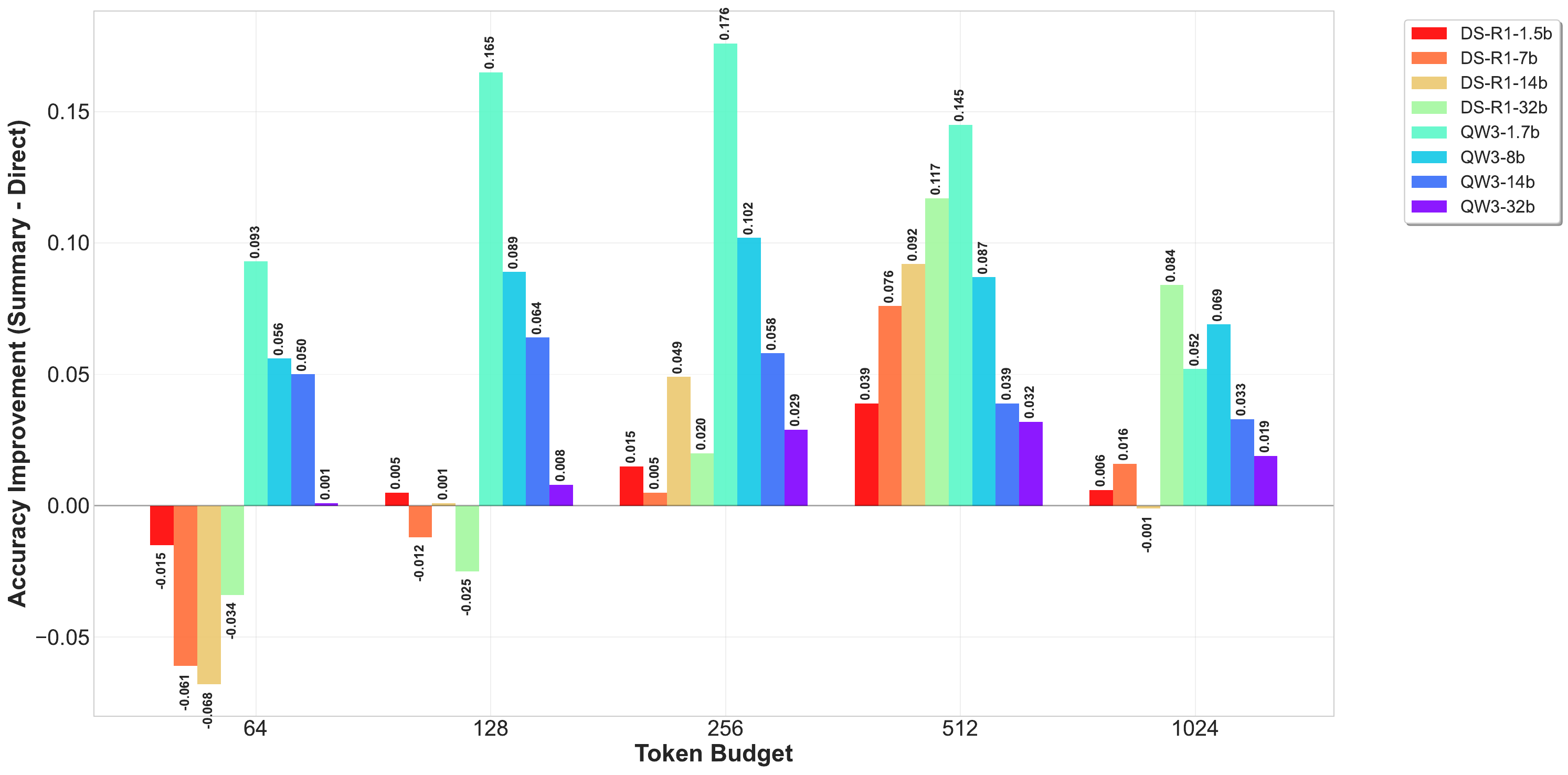}
\caption{English dataset performance gain across model configurations. The heatmap highlights where adaptive summarization improves over direct truncation, with the largest gains again concentrated in smaller answerers and tighter token budgets.}
\label{fig:en_improvement_heatmap}
\end{figure}

The English performance analysis (Figure~\ref{fig:en_improvement_heatmap}) displays consistent cross-model patterns, further confirming the reproducibility of language-independent reasoning transfer effects.

\begin{figure}[!ht]
\centering
\includegraphics[width=\columnwidth]{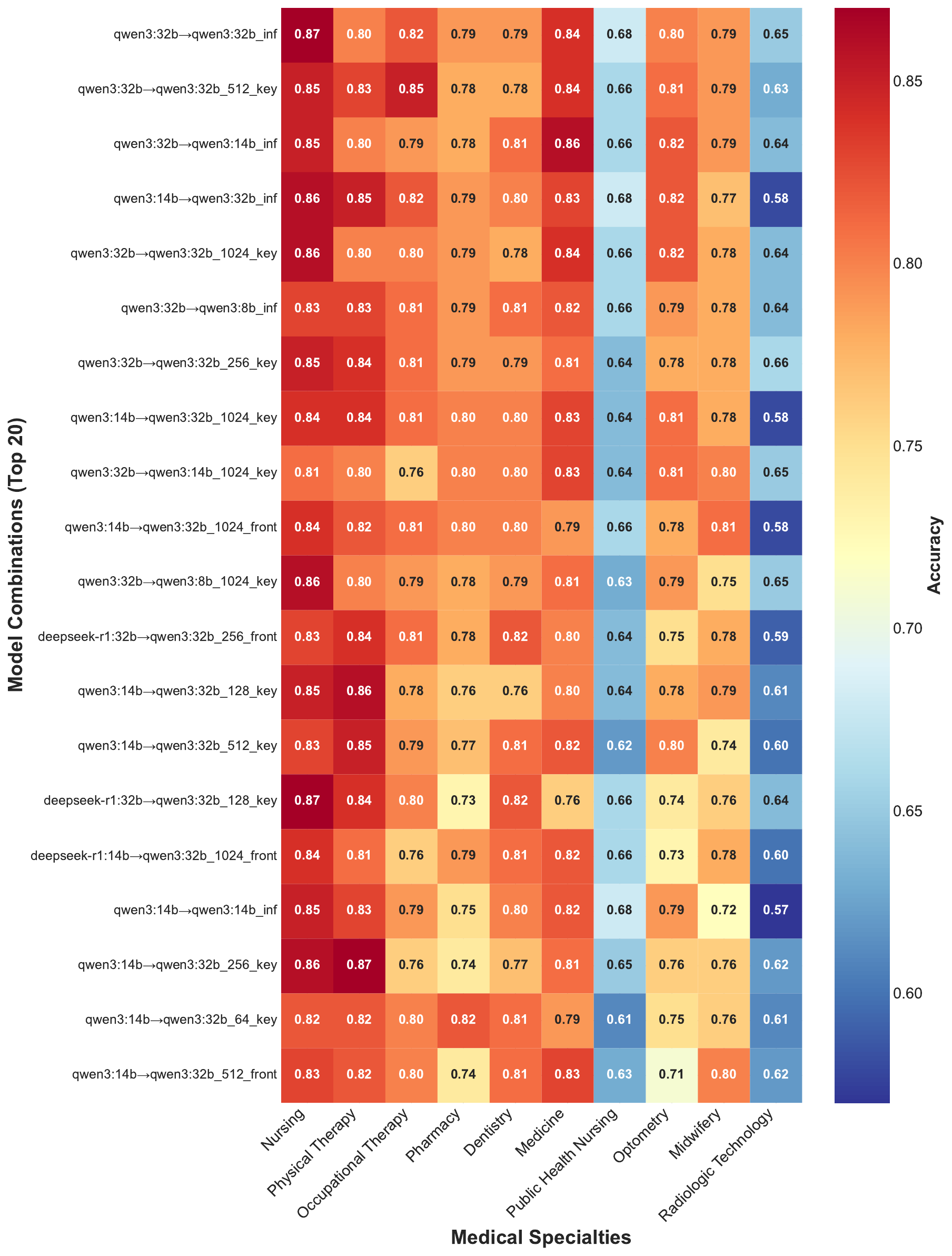}
\caption{English dataset top-configuration heatmap by medical specialty. Darker cells indicate stronger accuracy, and rows with consistently dark cells identify model pairs that retain robustness after translation.}
\label{fig:en_top_combinations}
\end{figure}

The model combination analysis (Figure~\ref{fig:en_top_combinations}) demonstrates similar inter-specialty trends to those found in the Chinese and Japanese datasets, indicating that model selection strategies generalize effectively across linguistic contexts.

\subsubsection{Comparative Token Distribution Analysis}

\begin{figure}[!ht]
\centering
\includegraphics[width=\columnwidth]{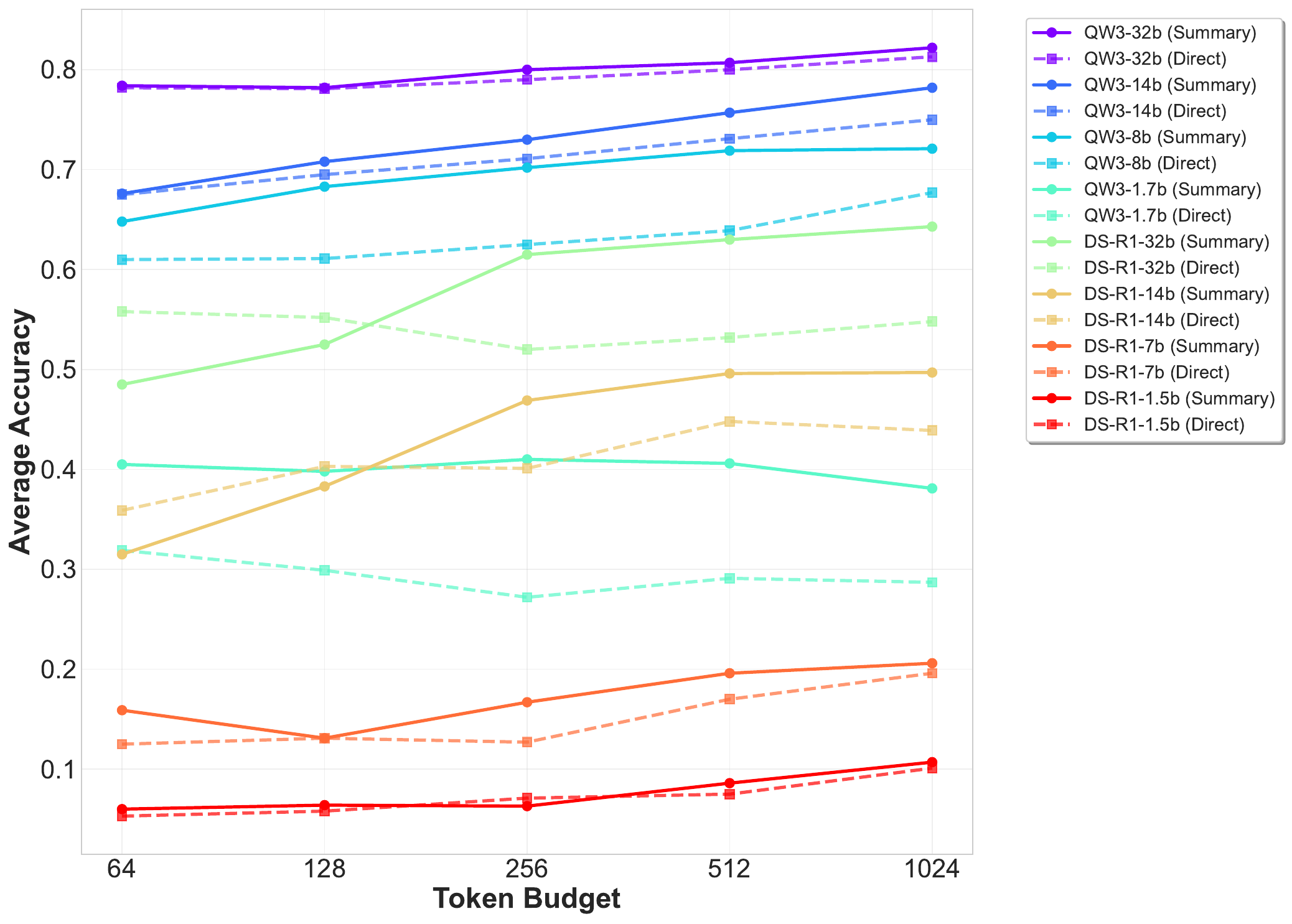}
\caption{Chinese dataset efficiency curves across token budgets. Solid trajectories reach their plateau earlier than truncation-based alternatives, indicating that CoT-X uses additional tokens more efficiently.}
\label{fig:cn_efficiency_curves}
\end{figure}

\begin{figure}[!ht]
\centering
\includegraphics[width=\columnwidth]{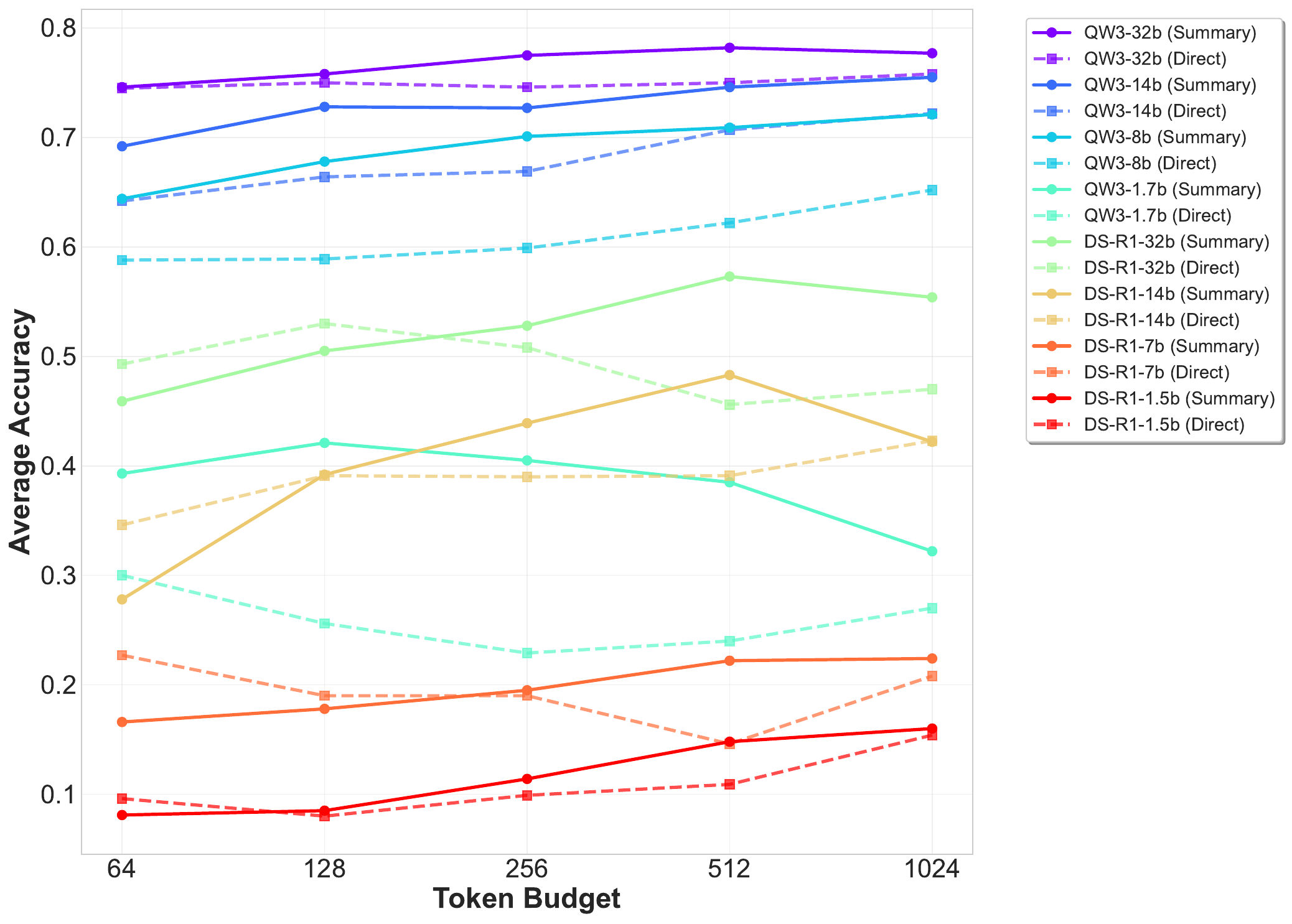}
\caption{English dataset efficiency curves across token budgets. The same saturation pattern observed in Japanese and Chinese appears here, supporting the language-agnostic behavior of adaptive CoT compression.}
\label{fig:en_efficiency_curves}
\end{figure}

The efficiency curves in Figures~\ref{fig:cn_efficiency_curves} and~\ref{fig:en_efficiency_curves} illustrate performance scaling with token budgets. All three languages exhibit consistent trends of improvement up to 1024 tokens, though with diminishing marginal gains at higher budgets.

\begin{figure*}[!ht]
\centering
\includegraphics[width=\textwidth]{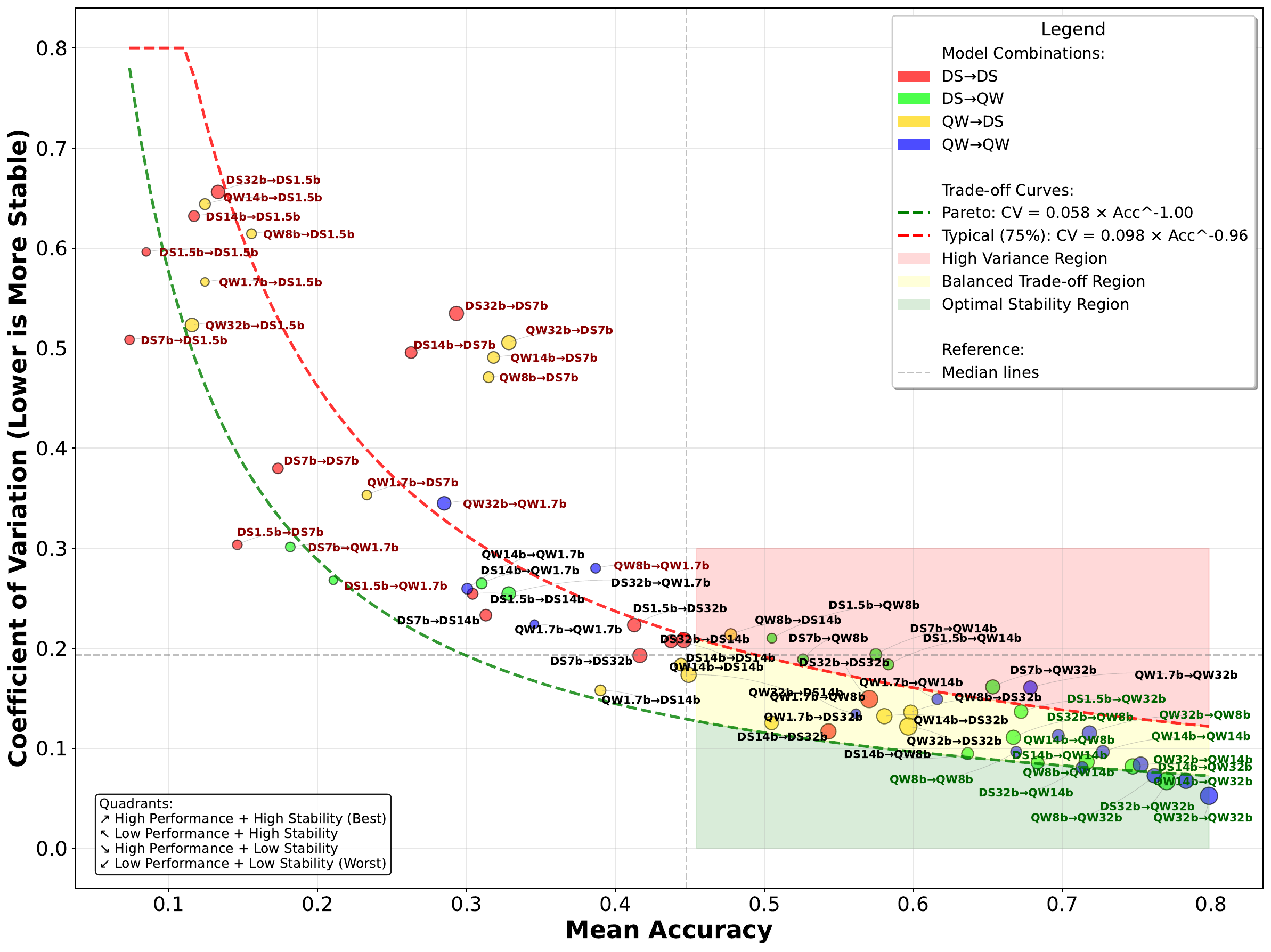}
\caption{Chinese dataset accuracy--robustness trade-off for $64$ model combinations. The x-axis reports mean accuracy and the y-axis reports coefficient of variation across specialties. The Pareto frontier identifies configurations that jointly preserve accuracy and stability.}
\label{fig:cn_pareto_frontier}
\end{figure*}

\begin{figure*}[!ht]
\centering
\includegraphics[width=\textwidth]{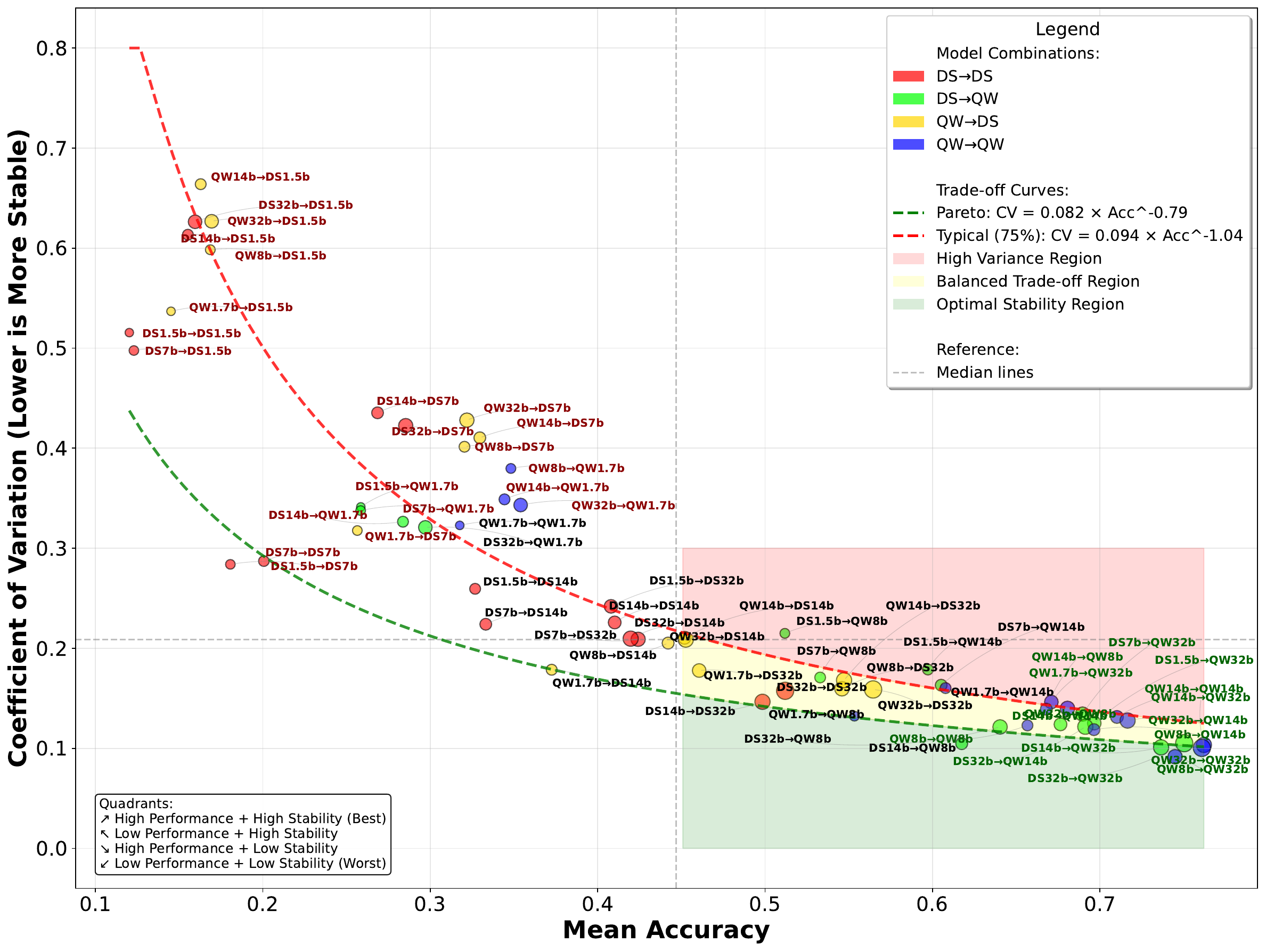}
\caption{English dataset accuracy--robustness trade-off for $64$ model combinations. The frontier closely follows the Japanese and Chinese patterns, suggesting that model-pair selection and adaptive compression remain useful across translated benchmarks.}
\label{fig:en_pareto_frontier}
\end{figure*}

The trade-off analysis between average accuracy and robustness (Figures~\ref{fig:cn_pareto_frontier} and~\ref{fig:en_pareto_frontier}) reveals a consistent power-law relationship across languages. Both Chinese and English datasets exhibit Pareto frontiers closely aligned with the Japanese benchmark, suggesting that the observed performance--stability trade-off is not specific to a single language version.

\subsubsection{Implementation Considerations for Multilingual Deployment}

Evaluations across the three datasets—Japanese, Chinese, and English—offer practical guidance for multilingual model deployment. Performance rankings remain stable across languages, with similar relative ordering among model architectures. Accuracy consistently improves with larger token budgets, particularly when increasing from 64 to 256–512 tokens.

Overall, the multilingual experiments confirm that the proposed chain-of-thought transfer framework maintains effectiveness across diverse linguistic contexts. The consistent trends in accuracy, robustness, and scalability underscore its suitability for global medical AI applications.

\end{document}